%% file: main.tex
\documentclass[acmtog]{acmart}

\usepackage{booktabs} 
\usepackage{times}
\usepackage{epsfig}
\usepackage{graphicx}
\usepackage{amsmath}
\usepackage{microtype}
\usepackage{comment}
\usepackage{float}
\usepackage{xcolor}
\usepackage{pifont}
\usepackage{animate}
\usepackage[utf8]{inputenc}
\usepackage{multicol}
\usepackage{multirow}
\usepackage{appendix}

\setcitestyle{square,nosort}

\newenvironment{tight_itemize}{
\begin{itemize}
  \setlength{\topsep}{0pt}
  \setlength{\itemsep}{0pt}
  \setlength{\parskip}{0pt}
  \setlength{\parsep}{0pt}
}{\end{itemize}}

\usepackage[ruled]{algorithm2e} 

\SetAlFnt{\small}
\SetAlCapFnt{\small}
\SetAlCapNameFnt{\small}
\SetAlCapHSkip{0pt}

\acmJournal{TOG}

\begin{document}
\title{Spatiotemporally Consistent HDR Indoor Lighting Estimation}

\author{Zhengqin Li}
\affiliation{
\institution{Meta Reality Labs Research, UC San Diego}
\country{USA}
}
\author{Li Yu}
\affiliation{
\institution{Meta Reality Labs}
\country{USA}
}
\author{Mikhail Okunev}
\affiliation{
\institution{Meta Reality Labs Research}
\country{USA}
}
\author{Manmohan Chandraker}
\affiliation{
\institution{UC San Diego}
\country{USA}
}
\author{Zhao Dong}
\affiliation{
\institution{Meta Reality Labs Research}
\country{USA}
}

\begin{abstract}
We propose a physically-motivated deep learning framework to solve a general version of the challenging indoor lighting estimation problem. Given a single LDR image with a depth map, our method predicts spatially consistent lighting at any given image position. Particularly, when the input is an LDR video sequence, our framework not only progressively refines the lighting prediction as it sees more regions, but also preserves temporal consistency by keeping the refinement smooth. Our framework reconstructs a spherical Gaussian lighting volume (SGLV) through a tailored 3D encoder-decoder, which enables spatially consistent lighting prediction through volume ray tracing, a hybrid blending network for detailed environment maps, an in-network Monte-Carlo rendering layer to enhance photorealism for virtual object insertion, and recurrent neural networks (RNN) to achieve temporally consistent lighting prediction with a video sequence as the input. For training, we significantly enhance the OpenRooms public dataset of photorealistic synthetic indoor scenes with around 360K HDR environment maps of much higher resolution and 38K video sequences, rendered with GPU-based path tracing. Experiments show that our framework achieves lighting prediction with higher quality compared to state-of-the-art single-image or video-based methods, leading to photorealistic AR applications such as object insertion. 

\end{abstract}
\begin{teaserfigure}
\animategraphics[width=\textwidth,poster=first]{2}{figures/teaser/frame}{00}{11}
\caption{We propose the first framework that achieves consistent high-quality indoor lighting prediction in both spatial and temporal domains. Our predicted HDR environment maps recover not only the visible and invisible light sources but also detailed reflection of visible surfaces, which enables realistic object insertion for both mirror and specular objects. Moreover, when the input is a video sequence, our framework can progressively refine our lighting prediction while keeping the transition smooth. (Please use Adobe Acrobat and click the figure to see the animated results.)}
\label{fig:teaser}
\end{teaserfigure}

\maketitle

\input{sources/introduction}
\input{sources/relatedworks}

\input{sources/volumerendering}
\input{sources/network}
\input{sources/experiment}
\input{sources/conclusion}

\bibliographystyle{ACM-Reference-Format}
\bibliography{ref}

\end{document}

%% file: sources/introduction.tex
\section{Introduction}
\label{sec:introduction}

\begin{figure*}
\centering
\animategraphics[width=\textwidth,poster=first]{2}{figures/pipeline/frame}{00}{12}
\caption{Overview of our deep learning-based framework to predict spatiotemporally consistent lighting at arbitrary locations of indoor scenes. Our framework consists of two parts: a spherical Gaussian lighting volume reconstruction network that reconstructs visible and invisible light sources and reflection, and a blending network that adds the detailed reflection from visible surfaces. Both parts use RNNs to accumulate observations from different frames and progressively refine our predictions. (Please use Adobe Acrobat and click the figure to see the animation.)}
\label{fig:pipeline}
\end{figure*}

The ubiquity of mobile devices for acquiring, sharing and editing videos suggests novel AR applications such as photorealistic scene enhancement and telepresence. High-quality consistent lighting estimation for a single image or across a video sequence is a crucial need for such applications. However, it remains a significant challenge, especially in indoor scenes where geometry, materials and light sources are diverse and produce images through complex, long-range interactions.

On the one hand, images or videos captured by the camera on mobile phones or tablets are usually very sparse and incomplete. For example, in a typical mobile phone photo, only $6\%$ of the panoramic scene is captured by the camera~\cite{legendre2019deeplight}. As a result, the inputs for lighting estimation are just low dynamic range (LDR) images with a very limited field of view (FoV). Therefore, to generate highly accurate indoor lighting estimation, one must hallucinate important HDR information and invisible parts of the scenes.

On the other hand, to allow photorealistic applications, the lighting prediction should satisfy several demands on quality. High-frequency directional lighting is essential for rendering high-quality shadows and specular highlights in scenarios such as bright sunlight through an open window. HDR lighting is necessary for rendering general non-mirror materials, while relatively high-resolution details need to be preserved for rendering realistic mirror reflections. Moreover, when rendering multiple (moving) objects, the predicted lighting should remain spatially consistent, which requires proper modeling of complex interactions between light sources, geometry, and materials. With a video sequence as input, it is advantageous to progressively improve the lighting prediction as more regions become visible while maintaining temporal consistency to avoid flickering. 

While several recent works propose deep learning-based frameworks for indoor lighting, they do not satisfy all the above requirements. Some prior works only achieve photorealistic outputs under certain circumstances (e.g. non-mirror surface \cite{li2020inverse,garon2019fast}, mirror surface only \cite{srinivasan2020lighthouse}
, diffuse lighting \cite{srinivasan2020lighthouse}, single object \cite{gardner2017indoor} etc.) or need extra non-trivial inputs (e.g. stereo image \cite{srinivasan2020lighthouse}). Further, none
of the existing methods take a video as input to progressively improve lighting prediction while preserving spatiotemporal consistency.

Motivated by the above, we propose a novel hybrid learning-based framework for consistent HDR indoor lighting prediction, taking either a single LDR image or video sequence as input. Our framework is designed to generate high-quality lighting prediction that enables various AR applications with photorealistic visual appearances. Table~\ref{tab:summary} compares recent lighting estimation methods, which shows that our method achieves all the desirable properties.

We achieve the above distinctions through the novel design of both physically-based representations and deep networks. The whole pipeline is summarized in Figure \ref{fig:pipeline}. We first reconstruct a spherical Gaussian lighting volume (SGLV) representation that is specifically designed to model complex HDR indoor lighting (Figure \ref{fig:SGLV_Definition}). Compared to the RGB$\alpha$ volume representation \cite{srinivasan2020lighthouse}, our representation inherits the advantages of rendering spatially consistent lighting while better modeling high-frequency directional lighting, like sunlight shining through windows. However, volume representations have difficulties recovering details of reflection, which are essential to render realistic mirror surfaces. Prior works increase the volume resolution or using multi-scale volumes \cite{srinivasan2020lighthouse,wang2021learning}, which makes their framework memory intensive and not suitable for mobile applications. Instead, we adopt a hybrid method by proposing a blending network to enhance details for the visible regions of the HDR environment map, which is lightweight and achieves higher quality reflection compared to prior works. To boost HDR lighting prediction and photorealism for AR applications, we introduce a physically-based rendering loss through an in-network Monte Carlo rendering layer that renders a glossy sphere from the lighting predictions. Finally, to handle video inputs, RNNs are introduced to progressively refine the lighting prediction while maintaining temporal consistency, even when the input depth maps are not fully consistent. 

Our framework is trained using the OpenRooms dataset \cite{li2020openrooms}, a large-scale, photorealistic synthetic indoor scene dataset with physics-based materials and lighting. The OpenRooms dataset contains over 120K HDR images with densely sampled spatially-varying per-pixel environment maps. We augment the dataset by rendering over 360K HDR environment maps at a much higher resolution of $120\times 240$ (compared to the original $16\times 32$), which enables us to render much more realistic reflection for mirror surfaces. To train our network for video inputs, we also render around 38K video sequences of $31$ frames each at a resolution of $240\times320$. Experiments on a widely-used real spatially-varying indoor lighting dataset show that our single image lighting estimation quality is comparable to the state-of-the-art \cite{li2020inverse}, while it can handle more complex materials with better spatial consistency. More importantly, to our best knowledge, our method is the first learning-based indoor lighting prediction framework that can enhance lighting prediction with video inputs and produce temporally consistent outputs. Figure \ref{fig:teaser} presents an animation of our spatiotemporally consistent HDR indoor lighting prediction, where we show 10 frames of our prediction from a video sequence. We include the full video in the supplementary to better demonstrate the quality of our lighting prediction. From the animation, we can see that our framework can recover both the detailed reflection and HDR invisible light sources, with spatial consistency as we insert mirror spheres at different locations. As we move our camera around, our method can progressively improve the prediction by adding more details and meanwhile, keep the transition smooth, as demonstrated in the reflection, highlights, and shadows of inserted objects.

In summary, our key contributions are:
\begin{tight_itemize}
\item A hybrid framework specifically designed for complex HDR indoor lighting, especially HDR light sources, directional lighting and detailed reflection.
\item RNN-based modules that progressively refine lighting prediction with video inputs, while maintaining temporal consistency even with not fully consistent depth inputs. 
\item State-of-the-art lighting prediction on real data, allowing photorealistic AR applications with fewer constraints.
\end{tight_itemize}

\begin{table}[t]
\centering
\includegraphics[width=\columnwidth]{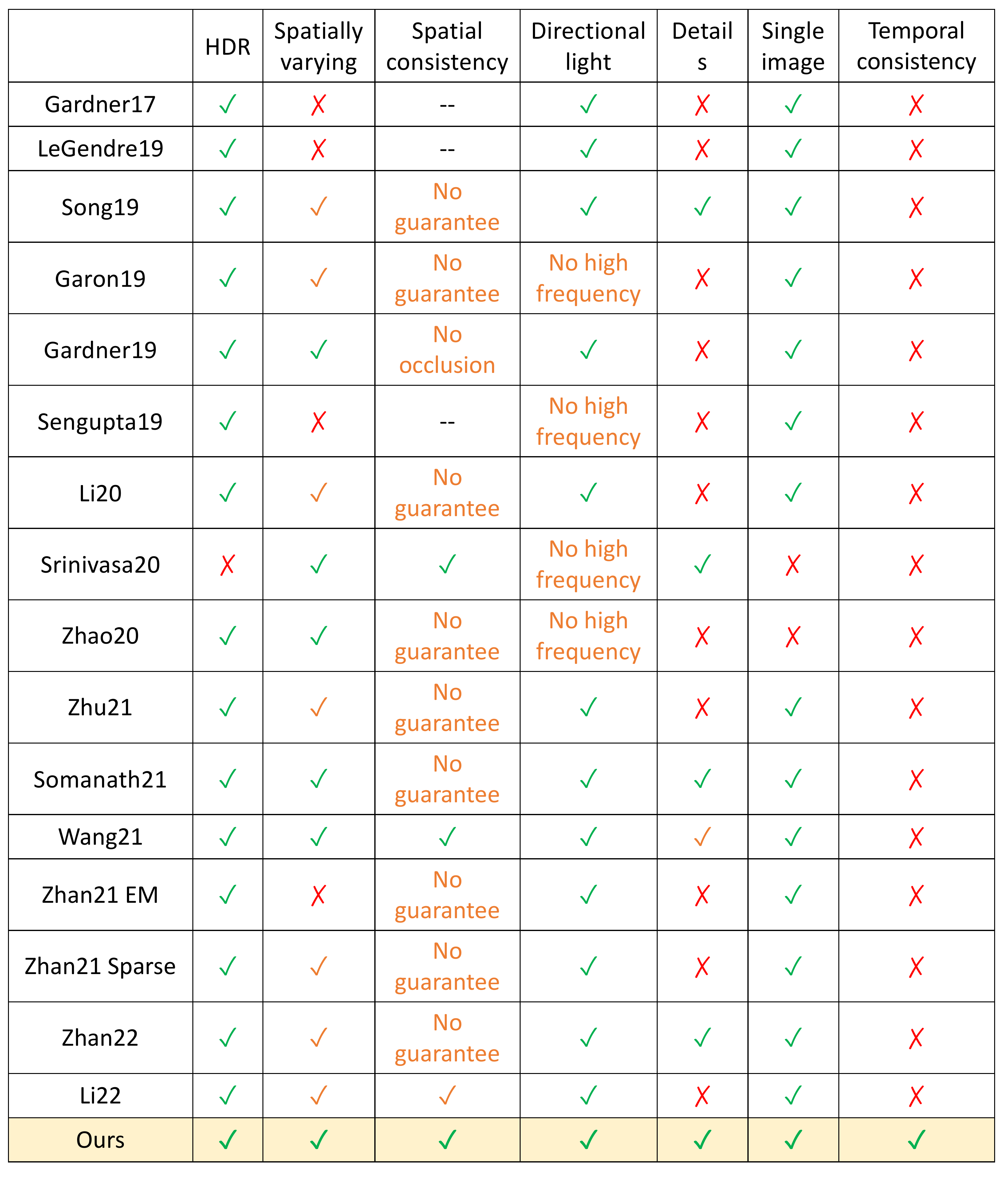}
\caption{A summary of state-of-the-art learning-based lighting estimation methods for indoor scenes. Our method achieves all the desirable properties and is the only method that can utilize video inputs to progressively refine predicted lighting while maintaining temporal consistency. }
\vspace{-0.6cm}
\label{tab:summary}
\end{table}

%% file: sources/relatedworks.tex
\section{Related Works}
\label{sec:relatedworks}

Indoor lighting estimation is a long-standing challenge in computer vision and graphics and is of fundamental importance to achieving photorealistic augmented reality. The pioneering work by Debevec captures indoor lighting by taking multiple images of a mirror sphere with different exposure times \cite{debevec1998rendering}.  While it can capture high-quality detailed HDR environment maps, it needs careful calibration, which is not friendly to non-expert users and cannot model spatially-varying lighting effects. Several model-based methods formulate indoor lighting estimation as part of a holistic inverse rendering framework and jointly reason about geometry, reflectance and lighting by minimizing hand-crafted energy functions. Notably, Barron and Malik \cite{barron2013intrinsic} model the indoor scene appearance with a diffuse reflectance map and a mixture of geometry and lighting basis driven from data priors, which allows them to simultaneously recover diffuse reflectance, refined depth map and per-pixel lighting from a single RGBD image. Karsch et al. \cite{karsch-tog-14} approximate indoor lighting by selecting a HDR panorama from a large-scale dataset. Then, they optimize the light source position and intensity by solving an inverse rendering problem based on estimated diffuse reflectance and depth map. These methods rely on certain assumptions to simplify the problem and therefore may not achieve high-quality lighting estimation in complex scenes, with the presence of complex directional lighting, strong shadows caused by large occlusion and highly glossy materials. 

Recent methods apply deep learning frameworks to tackle this highly ill-posed classical challenge. However, as summarized in Table \ref{tab:summary}, none of them can achieve all the desirable properties for indoor lighting estimation. Akimoto et al. \cite{akimoto2022diverse} and Dastjerdi et al. \cite{dastjerdi2022guided} extrapolate a 360 degree field of view panorama from a single image at the center of it using a deep generative network. Gardner et al. \cite{gardner2017indoor}, Sengupta et al. \cite{sengupta2019neural}, Zhan et al. \cite{zhan2021emlight}, and LeGendre et al. \cite{legendre2019deeplight} predict a single lighting for the whole indoor scene. These methods cannot model important spatially-varying lighting. Several methods can predict spatially-varying lighting from a single indoor image. Particularly, Garon et al. \cite{garon2019fast} and Zhao et al. \cite{zhao2020pointar} use spherical harmonics bases \cite{ramamoorthi2001efficient} to approximate indoor HDR environment maps and therefore can only recover low frequency signals. Li et al. \cite{li2020inverse}, Zhu et al. \cite{zhu2022irisformer}, and Li et al. \cite{li2022physically} use spherical Gaussian lobes while Zhan et al. \cite{zhan2021sparse} use wavelets instead to better preserve high-frequency directional lighting. However, neither of them can recover the near-field details in an environment map, which is important for rendering realistic reflectance on mirror and highly glossy surfaces. Song et al. \cite{Song_2019_CVPR}, Somanath et al. \cite{somanath2021hdr}, and Zhan et al. \cite{zhan2022gmlight} directly predict the 2D environment map by inpainting the partial observation from the input image through generative modeling. Nevertheless, one important limitation of all the above spatially varying lighting prediction methods is that they cannot guarantee spatial consistency. This may cause flickering artifacts if we use their predicted lighting to render a virtual object moving around the scene. In contrast, Gardner et al. \cite{gardner2019parametric} and Srinivasan et al. \cite{srinivasan2020lighthouse} can predict spatially consistent lighting at any positions,  by reconstructing environment maps from their 3D scene representations, i.e. spherical Gaussian light sources and RGB$\alpha$ volume, through physics-based ray tracing. However, Gardner et al. cannot handle occlusion, while Srinivasan et al. cannot handle HDR and directional lighting. A very recent work \cite{wang2021learning} by Wang et al. solves the limitation of \cite{srinivasan2020lighthouse} by introducing spherical Gaussian volume to handle directional lighting, which is similar to our proposal. However, both of them use high-resolution volumes to recover detailed reflection while our hybrid representation achieves better details with much less computational cost. More importantly, none of them can handle video inputs with temporal consistency.

The only prior method that considers video inputs is proposed by Wei et al. \cite{wei2020object}. However, their method is not demonstrated to be able to use incoming frames to progressively improve the lighting prediction. On the contrary, our method is the first learning-based indoor lighting prediction framework that can enhance lighting prediction with video input, while preserving both spatial and temporal consistency. Our method achieves all the desirable properties for indoor lighting prediction, which allows us to obtain a higher level of photorealism in various AR applications with better generalization.

%% file: sources/volumerendering.tex
\begin{figure}[t]
	\centering
	\includegraphics[width=\columnwidth]{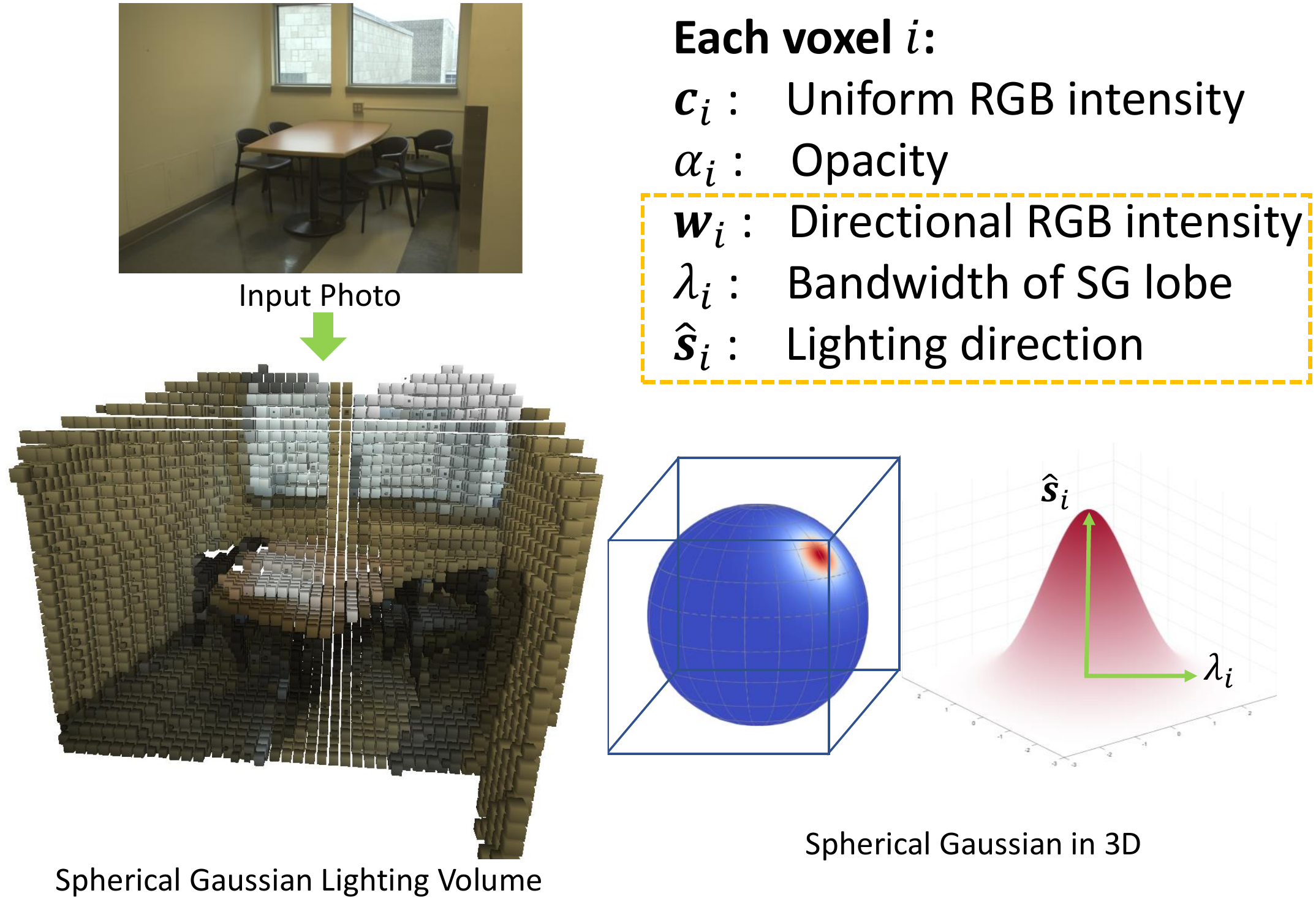}
	\caption{Visualization of spherical Gaussian lighting volume.}
	\label{fig:SGLV_Definition}
\end{figure}

\section{Spherical Gaussian Lighting Volume}
\label{sec:volumerendering}

Indoor lighting presents the most complex and diverse challenges for lighting estimation, such as complex occlusion, directional sunlight, and strong global illumination. Thus, we need a lighting representation that is expressive enough to model various types of light transport. Further, this representation should enable fusing the observations from different views, to allow extension to video inputs (\S~\ref{sec:network}).

We therefore adopt a volumetric lighting representation to render an HDR environment map at any given location through differentiable volume ray tracing, which allows spatially consistent lighting prediction with details, while enabling multi-view fusion for video inputs. The starting point is the representation of \cite{srinivasan2020lighthouse}, which defines an RGB color and an opacity $\alpha$ for each voxel. Such a representation assumes each voxel uniformly emits light in every direction, hence, poorly deals with directional lighting, such as the sunlight through the windows. Following the success of the recent work \cite{li2020inverse}, we enhance the volumetric representation by introducing an additional spherical Gaussian lobe to better model directional lighting. We term our representation as spherical Gaussian lighting volume (SGLV). Similar representation is proposed in a recent work by Wang et al. \cite{wang2021learning}.

Our SGLV representation is illustrated in Figure~\ref{fig:SGLV_Definition}. Besides an RGB$\alpha$ value, each voxel is augmented with three SG parameters: $\mathbf{w} \in {\mathbb{R}}^3$ to control the intensity, $\lambda  \in {\mathbb{R}}$ to control the bandwidth and unit vector $\hat{\mathbf{s}} \in {\mathbb{R}}^3$ to control the direction.  When rendering an HDR environment map using SGLV, we modify the volume ray tracing method accordingly to accumulate both the RGB value $\mathbf{c}$ and the SG parameters $\{\mathbf{w}, \lambda, \hat{\mathbf{s}}\}$ using standard differentiable volume ray tracing. More specifically, suppose we render the HDR environment map $L$ at a randomly sampled location. Let $\mathbf{\hat{l}}$ be a ray corresponding to an arbitrary pixel of $L$ and $\mathcal{I}$ be the indices of points sampled uniformly on the ray $\mathbf{\hat{l}}$, organized in ascending order according to their distance to the sampled location. The accumulation of all the SGLV parameters for the ray $\mathbf{\hat{l}}$ can be computed as: 
\begin{equation}
\mathbf{x_{\hat{l}}} = \sum_{i\in \mathcal{I}} \alpha_{i}\mathbf{x}_{i} \prod_{j < i, j \in \mathcal{I}} (1-\alpha_{j}), \quad \mathbf{x} \in \{\mathbf{c, w}, \lambda, \hat{\mathbf{s}}\}.
\end{equation}
At each step point, both $\mathbf{x}_{i}$ and $\alpha_{i, j}$ are obtained through trilinear interpolation from nearby voxels so that the whole process is differentiable. After obtaining the accumulated parameters, we compute the intensity $L_{\mathbf{\hat{l}}}$ along direction $\mathbf{\hat{l}}$ of the HDR environment map $L$ as: 
\begin{equation}
L_{\mathbf{\hat{l}}} = \mathbf{c_{\hat{l}}} + \mathbf{w_{\hat{l}} }\exp\Big(\lambda(\mathbf{\hat{l}\cdot\hat{s}_{\hat{l}} } - 1) \Big)
\end{equation} 

\begin{figure}
\centering
\includegraphics[width=\columnwidth]{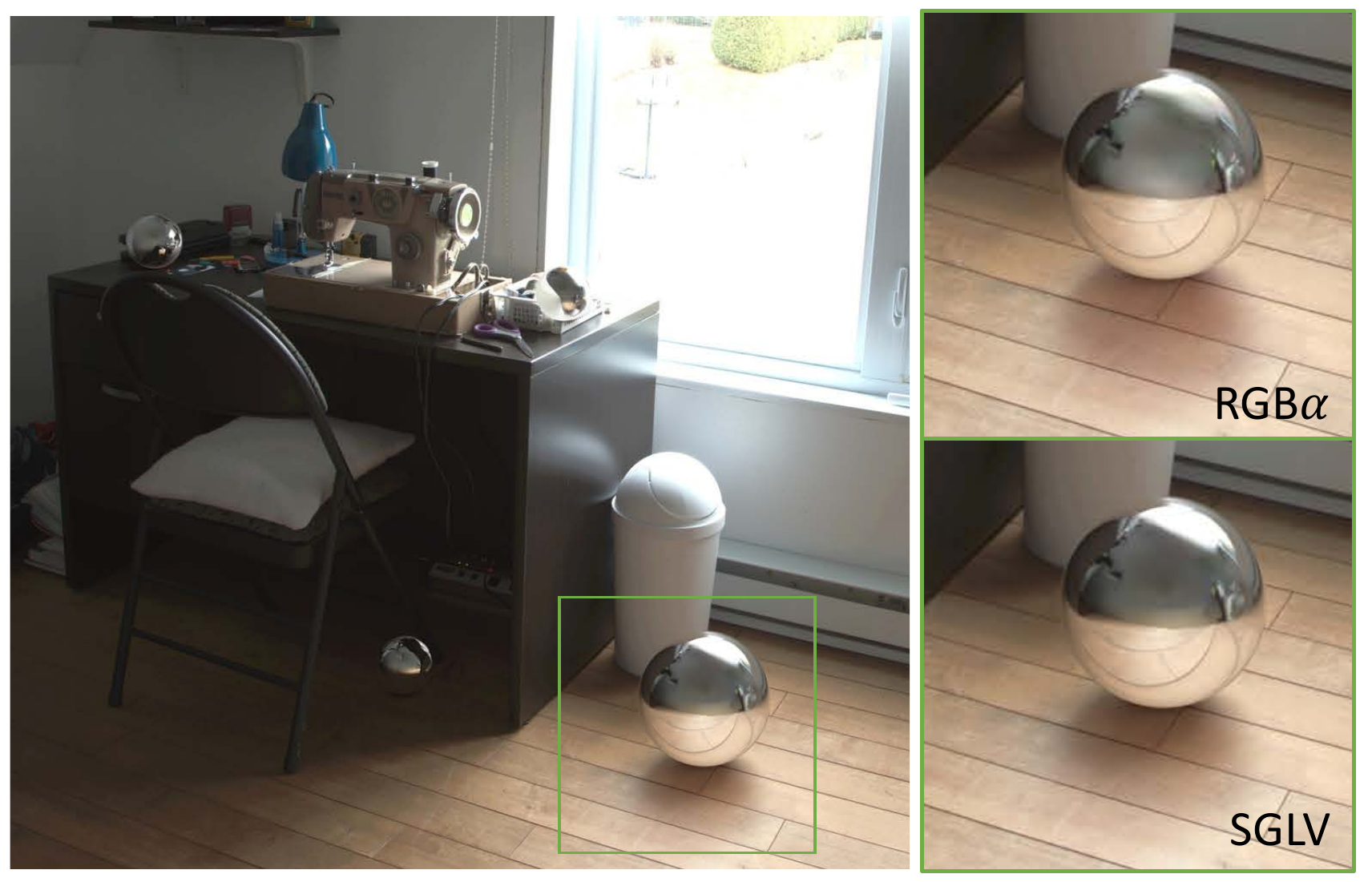}
\caption{Comparison of RGB$\alpha$ and our spherical Gaussian  lighting volume (SGLV) for indoor lighting estimation on a real example. The SGLV allows a better encoding of high-frequency directional lighting in indoor scenes, which leads to sharper highlights and shadows}
\label{fig:SGvsRGB}
\end{figure}

Figure \ref{fig:SGvsRGB} shows a comparison of lighting prediction between the  RGB$\alpha$ volume and our proposed SGLV. We train our proposed method with two different lighting representations and then use the lighting predictions from two different representations to render virtual mirror spheres into a real indoor scene image.  It clearly shows that our SGLV representation can better model the high-frequency directional sunlight coming from the window, leading to more realistic specular highlights and shadows. 

%% file: sources/network.tex
\section{Physically Motivated Network Design}
\label{sec:network}

\begin{figure*}
\includegraphics[width=\linewidth]{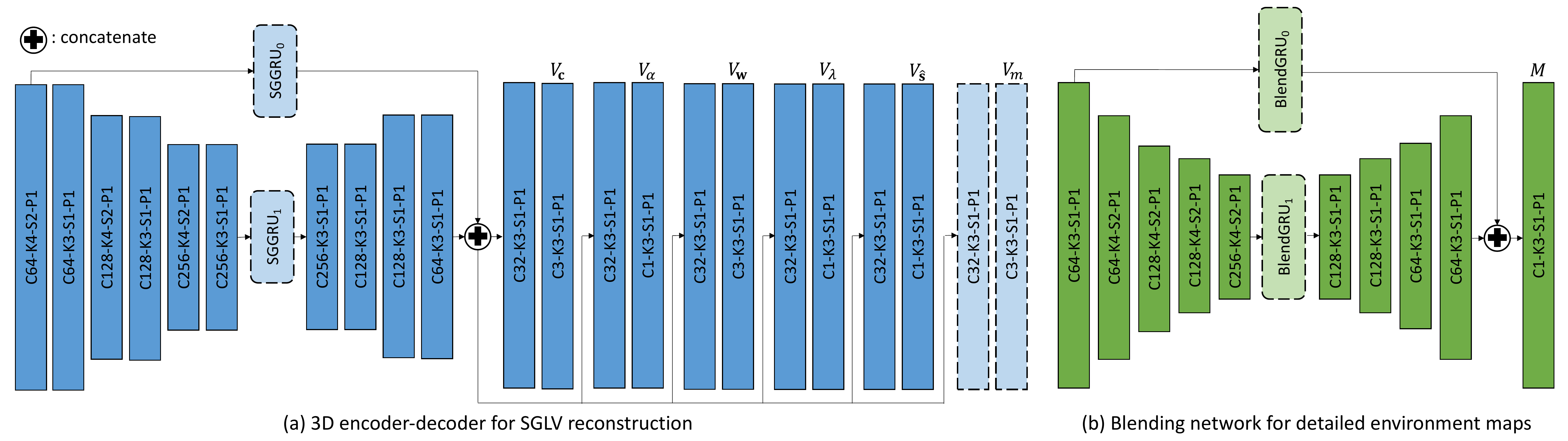}
\caption{Summary of our physically-motivated deep network architecture for indoor lighting estimation from a single image or a video sequence. $Ca$-$Kb$-$Sc$-$Pe$ means a convolution layer with channel $a$, kerneal size $b$, stride $c$ and padding $e$. The RNN modules added for handling video sequences are surrounded by black dashed lines. Specifically, our encoder-decoder architectures have only two skip-connections, which allows us to preserve high-frequency signals but also reduce memory consumption, especially for video inputs. }
\label{fig:network}
\end{figure*}

We now present our physically motivated deep learning based framework for spatially and temporally consistent high-quality HDR indoor lighting estimation at arbitrary locations. The overall architecture is summarized in Figure \ref{fig:network}. We first introduce our light-weight 3D CNN for single image SGLV prediction, followed by a 2D blending CNN to improve the details. Next, we extend our frameworks with RNN architectures so that we can handle video sequences with temporal consistency. At the end of this section, we discuss the dataset creation process and important implementation details. 

\subsection{Single image indoor lighting prediction}
\paragraph{3D volume initialization.} We first build an initial 3D RGB$\alpha$ volume as the input to our 3D CNN for volume completion, by projecting the input image $C$ into 3D space based on depth map $D$, which can either be predicted by a deep network or captured by a sensor. We initialize the $\alpha$ value first. Let $\mathbf{v}$ be the center of a voxel, $v$ be the voxel's side length, $\text{proj}(\cdot)$ be the projection function projecting to the image plane, $\mathbf{o}$ be the center of the image plane and $\mathbf{\hat{z}}$ be its orientation. We define $D_{\mathbf{p}}$ to be the depth value of $\mathbf{p}$ on the image plane, computed through bilinear interpolation. The initial $\alpha$ is computed as
\begin{eqnarray*}
\alpha_{\mathbf{v}} &=& \left\{
\begin{array}{cc}
4(\frac{1}{v}( (\mathbf{v-o})\cdot\mathbf{\hat{z}} - D_{\text{proj}(\mathbf{v})} ) + 1)    &  D_{\text{proj}(\mathbf{v})} > (\mathbf{v-o})\cdot\mathbf{\hat{z}} \\[8pt]
4(\frac{1}{v}(D_{\text{proj}(\mathbf{v})} - (\mathbf{v-o})\cdot\mathbf{\hat{z}}) + 5)    &  D_{\text{proj}(\mathbf{v})} \leq (\mathbf{v-o})\cdot\mathbf{\hat{z}}
\end{array}
\right. \\[8pt]
\alpha_{\mathbf{v}} &=& \text{clip}(\alpha_\mathbf{v}, 0, 1)
\end{eqnarray*}
Intuitively, we smooth the $\alpha$ value in front of and behind the scene surface unevenly so that the rendered HDR environment map can have sharp edges but without holes. For voxels that are behind the camera or outside the camera frustum, we set their initial $\alpha$ to be 0. 

Given the value of $\alpha$, the initial color value $\mathbf{c}$ is computed as 
\begin{equation*}
\mathbf{c}_{\mathbf{v}} = \alpha_{\mathbf{v}}C_{\text{proj}(\mathbf{v})},
\end{equation*}
where $C_{\text{proj}(\mathbf{v})}$ is similarly computed through bilinear interpolation. However, this initial RGB$\alpha$ volume does not yet encode all the information from the depth map. Namely, the space inside the camera frustum, between the camera and the scene surface, should be empty. To incorporate this information, we augment the initial volume with another channel $e$. We set $e_\mathbf{v}$ to be $-1$ if $\mathbf{v}$ should be empty according to the depth map $D$:
\begin{equation*}
e_{\mathbf{v}} = \left\{
\begin{array}{cl}
-1    & \frac{1}{v}(D_{\text{proj}(\mathbf{v})} - (\mathbf{v-o})\cdot\mathbf{\hat{z}}) > 3  \\[8pt]
0     & \text{otherwise}.
\end{array}
\right.
\end{equation*}
As shown above, we only set $e_\mathbf{v}$ to be -1 if $\mathbf{v}$ is three voxels away from the scene surface, in case the depth map $D$ is inaccurate.

\paragraph{3D encoder-decoder for SGLV reconstruction} We train a relatively light-weight 3D encoder-decoder to reconstruct SGLV from the initial RGB$e\alpha$ volume. Formally, let $\tilde{V}$ represent the initial volumes and $V$ represent predicted volumes. Our 3D CNN encoder-decoder can be written as
\begin{eqnarray*}
V_{f_0}, V_{f_1} &=& \mathbf{SGEncoder}(\tilde{V}_{\mathbf{c}}, \tilde{V}_{\alpha}, \tilde{V}_{e} ) \\
V_{\mathbf{c}}, V_{\alpha}, V_{\mathbf{w}}, V_{\lambda}, V_{\hat{\mathbf{s}}} &=& \mathbf{SGDecoder}(V_{f_0}, V_{f_1})
\end{eqnarray*}
where $V_{f_0}$ and $V_{f_1}$ are two feature volumes connecting encoder and decoder, as shown in Figure \ref{fig:network}.  

It is important to ensure that the predicted volumes are consistent with the input depth map, i.e. the voxels inside the camera frustum in front of scene surfaces should always be empty. Otherwise, the lighting prediction near the surface can be wrong due to the occlusion of nearby voxels. Therefore, we explicitly add a constrain to clear these near-surface voxels.
\begin{equation}
V_{\mathbf{x}} = V_{\mathbf{x}} (1 + \tilde{V}_{e}), \quad \mathbf{x} \in \{\mathbf{c}, \alpha, \mathbf{w}, \lambda, \mathbf{\hat{s}}\}
\end{equation}

Unlike Lighthouse \cite{srinivasan2020lighthouse}, which uses a heavyweight multi-scale network that requires 16G GPU memory for training, we make our 3D CNN network much more light-weight so that it can be trained on a 1080Ti GPU. Three design choices are specifically made. First, we set the number of voxels to be only one fourth of that of Lighthouse, as detailed in Section \ref{sec:details}. While this will cause more blurry predictions, we solve it by introducing a cheap 2D blending module, which will be discussed later. Further, as shown in Figure \ref{fig:network}, we keep only two connections $V_{f_0}$ and $V_{f_1}$ for our encoder-decoder architecture, which connect only the first and the last layers from the encoder to the decoder. For single image inputs, this not only allows us to keep reasonable details but also can reduce the GPU memory consumption by 30\% compared to a standard U-net. For video inputs, we have only two levels of features to be updated by RNNs, which greatly reduces the computational cost for both training and testing. Finally, as shown in Figure \ref{fig:network}, in our 3D decoder, the branches to output the spherical Gaussian volumes $V_{\mathbf{w}}$,  $V_{\hat{\mathbf{s}}}$, $V_{\lambda}$ and the weight volume $V_{m}$ share features with the branches to output $V_{\mathbf{c}}$ and $V_{\alpha}$. Therefore, the computational overhead to have the extra outputs is very small compared to that of RGB$\alpha$ volume.

\begin{figure*}
\centering
\includegraphics[width=\textwidth]{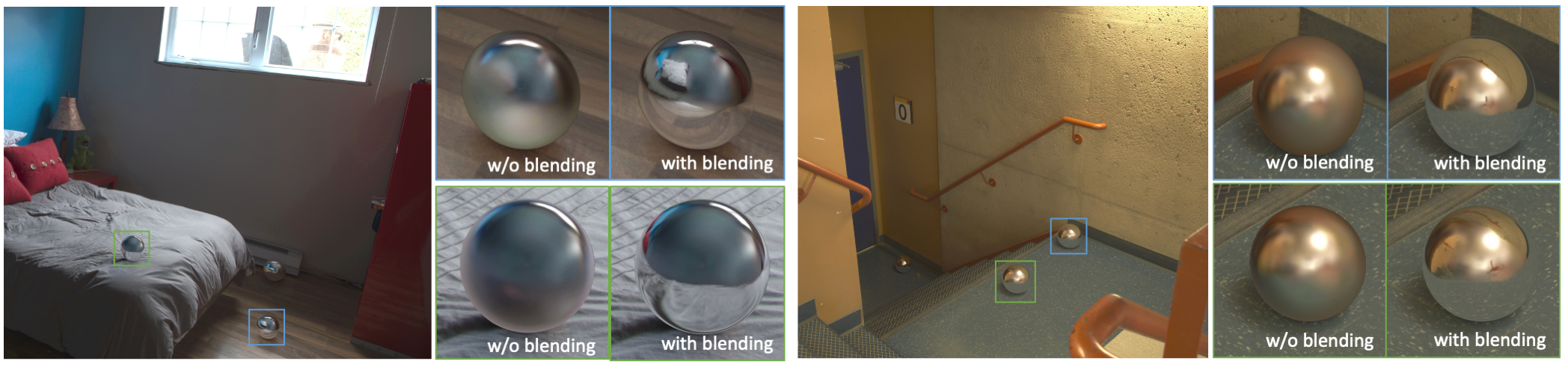}
\caption{Comparison of lighting prediction with and without the blending module on mirror sphere insertion. Our blending module adds significantly amount of details to render photorealistic reflection, while keeping the computational overhead significantly lower than prior methods \cite{srinivasan2020lighthouse, wang2021learning}. }
\label{fig:blending}
\end{figure*}

\paragraph{Blending network for detailed environment maps} A high-resolution detailed environment map is necessary to render photorealistic reflection when we insert a highly glossy object into the scene. However, to directly generate such an environment map through volume ray tracing, we need to either reconstruct a high-resolution volume \cite{wang2021learning} or a multi-scale volume \cite{srinivasan2020lighthouse} by allocating denser voxels in the camera frustum, where the virtual object will be inserted. Both still lead to high computational cost. On the contrary, we solve this issue in a simple but effective hybrid manner. Given the depth map $D$, we build a partial mesh of the indoor scene by first projecting pixels into 3D space and then connecting adjacent pixels. We then texture the partial mesh with the input image and render a detailed LDR partial environment map efficiently with a ray tracer. This detailed LDR partial environment map complements the predicted HDR environment map but directly blending them together can cause the loss of HDR lighting and edge artifacts. Therefore, we train a light-weight 2D blending network that can output a blending weight. Formally, let $\dot{L}$ be the HDR complete environment map rendered from SGLV. With some abuse of notation, let $\tilde{L}$ be the detailed LDR partial environment maps and $\tilde{L}_{M}$ be the binary mask indicating the visible regions, both rendered from the partial mesh. Our blending network is defined as
\begin{eqnarray*}
F_{0}, F_{1} &=& \mathbf{BlendEncoder}(\dot{L}, \tilde{L}, \tilde{L}_{M}, 1-\tilde{L}_{M}) \\
L_M &=& \mathbf{BlendDecoder}(F_0, F_1) \\
L &=& \dot{L}(1-L_M) + \tilde{L} L_M
\end{eqnarray*}
where $F_0$ and $F_1$ are two 2D feature maps connecting encoder and decoder. $L_M$ is a single channel blending weight in the range [0, 1]. Similar to our 3D encoder-decoder, our 2D blending encoder-decoder also only keeps 2 connections to reduce the computational cost, especially when the input is a video sequence. Figure \ref{fig:blending} compares the environment map predictions directly rendered from SGLV and from our blending network, which clearly shows the improved details and more realistic reflection when using it to render mirror sphere. While our blending model can only recover details of visible surfaces of the scene, we argue that hallucinating details of invisible surfaces is a too challenging problem as prior methods using larger volume trained with adversarial loss cannot handle this issue satisfactorily either. Figure \ref{fig:videoSyn} and \ref{fig:videoReal} further compare our lighting prediction with \cite{srinivasan2020lighthouse}. We observe that on both synthetic and real data, our method achieves higher or similar level of details for visible parts of the scenes but with much less cost. For invisible parts, our method can better recover the overall color of reflection and most importantly the HDR light sources, leading to much more realistic object insertion results with both mirror and glossy materials. 

\begin{figure}
\centering
\includegraphics[width=\columnwidth]{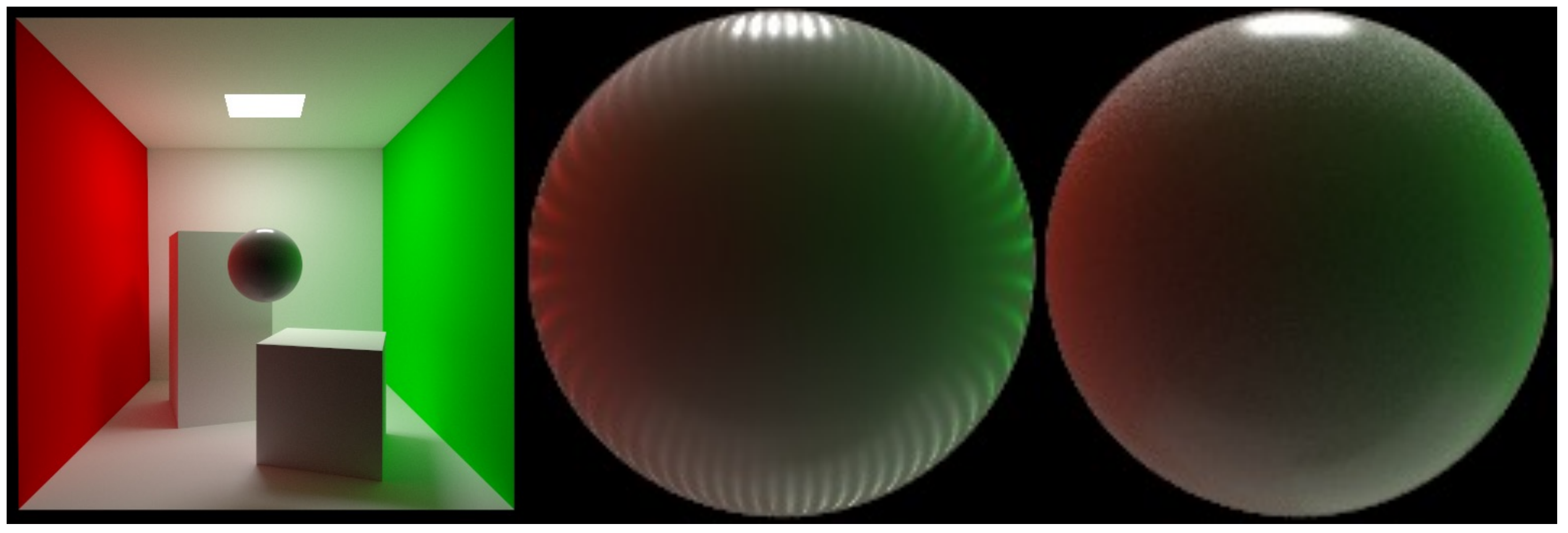}
\caption{Comparisons between rendering layers that uniformly samples the $\theta$-$\phi$ space and that does importance sampling according to the microfacet BRDF. From left to right: the cbox scene with a inserted glossy sphere, the inserted glossy sphere rendered by a rendering layer using uniform sampling, as in \cite{li2020inverse}, and importance sampling.}
\label{fig:renderingLayer}
\end{figure}

\paragraph{Monte Carlo in-network rendering layer.}
Since our eventual goal is to build high-quality AR applications such as virtual object insertion, we introduce a Monte Carlo in-network rendering layer to directly optimize the network for this purpose. Our rendering layer takes an HDR environment map and renders a virtual glossy sphere facing the camera. The reason we render a glossy sphere instead of a diffuse one is that the former can reflect specular highlights, providing a good cue for the network to recover the high-frequency directional lighting. In practice, we use the microfacet BRDF model proposed in \cite{brdfModel}, with its diffuse albedo set to be $(0.8, 0.8, 0.8)$ and roughness to be 0.2. During training, we render two spheres using predicted and ground truth lighting and compare the two spheres to compute a rendering loss, as shown in Eq. \eqref{eq:renderinLoss}. 

A similar rendering layer has also been used in the prior work \cite{li2020inverse} for indoor lighting estimation. However, their rendering layer integrates the incoming radiance by uniformly sampling the $\theta$-$\phi$ space, which is not optimal and can cause aliasing artifacts. On the contrary, our rendering layer samples the hemisphere through importance sampling according to the microfacet BRDF model, which leads to  less noise and more realistic specular highlights. Let $R$ be the rendered image, $\Omega$ be the set of directions sampled according to the microfacet BRDF and $P(\cdot)$ be the sampling probability. We have
\begin{equation*}
R_{\mathbf{p}} = \sum_{\mathbf{\hat{\mathbf{l}}}\in \Omega_{\mathbf{p}}} \frac{\text{F}(\mathbf{\hat{l}}, \mathbf{\hat{n}_p}, \mathbf{\hat{v}_p}) L_{\hat{\mathbf{l}}}}{\text{P}_{\mathbf{p}}(\mathbf{\hat{l}} ) },
\end{equation*}
where $\text{F}(\cdot)$ is the microfacet BRDF. $\hat{\mathbf{n}}_{\mathbf{p}}$ and $\hat{\mathbf{v}}_{\mathbf{p}}$  are the normal and view direction for pixel $\mathbf{p}$. Figure \ref{fig:renderingLayer} compares the rendered images using the two sampling strategies on the Cornell box scene. With the same number of samples, uniform sampling results in aliasing artifacts, while our importance sampling can render smooth appearance.

\begin{figure}
\centering
\includegraphics[width=\columnwidth]{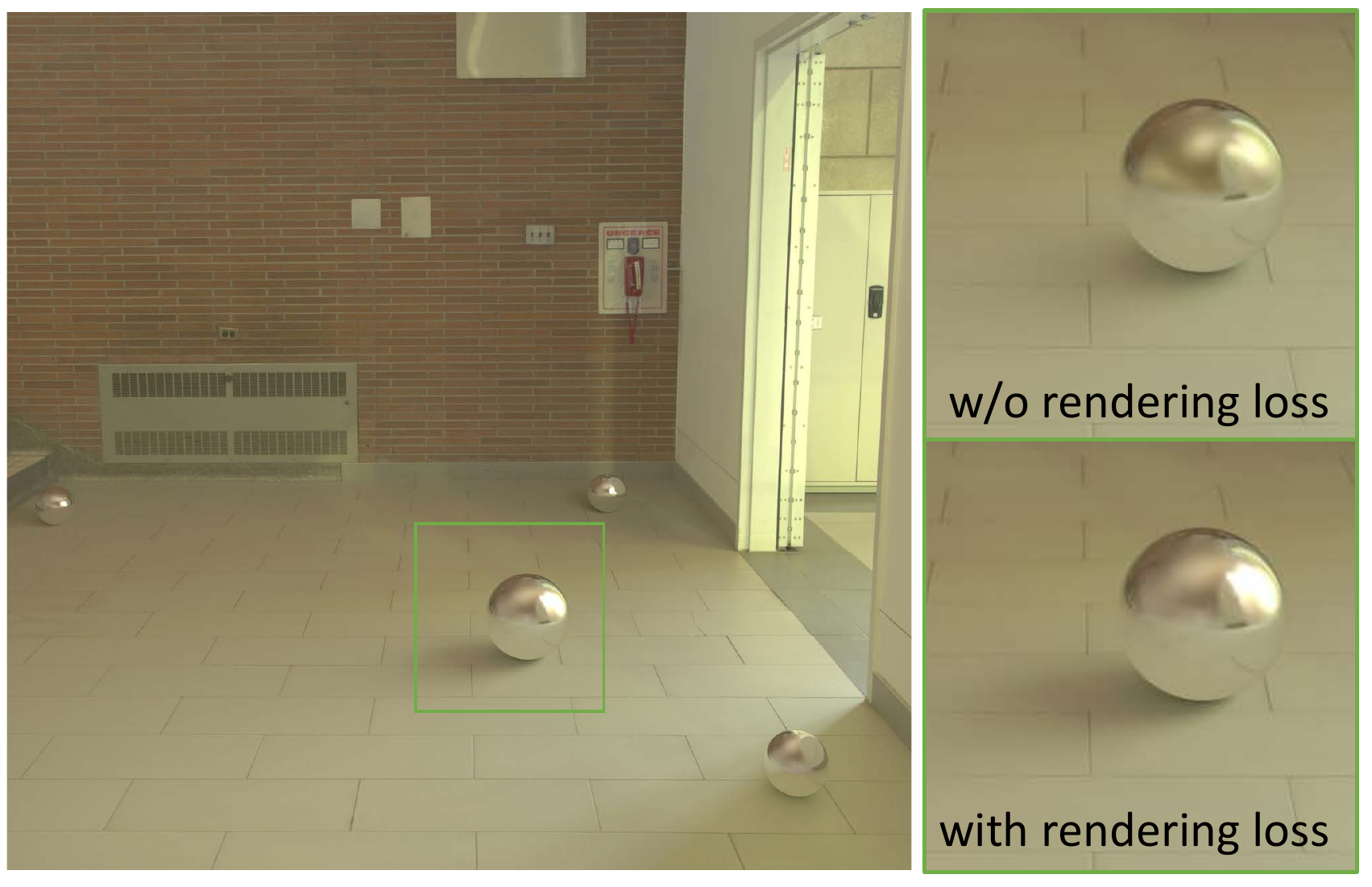}
\caption{Comparison of indoor lighting estimation with and without rendering loss on a real example. Rendering loss encourages the network to better recover HDR light sources.}
\label{fig:renderingLayerObj}
\end{figure}

\paragraph{Loss functions.} We use two loss functions for single view indoor lighting estimation: a per-pixel lighting loss and a rendering loss. The per-pixel lighting loss is defined as the $\log$ $L_2$ loss between our predicted HDR environment maps and the ground truth, located at three randomly sampled positions inside the camera frustum.
\begin{equation}
\mathcal{L}_{L_2} = \sum_{\{L, \bar{L}\}}\Big(\log(L + 1) - \log(\bar{L} + 1) \Big)^2,
\label{eq:logl2}
\end{equation}
where $\bar{L}$ is a ground truth HDR environment map. As for the rendering loss, we clamp the images $R$ rendered with our lighting prediction and the ground truth to the range $[0, 1]$, then compute their $L_2$ distance:
\begin{equation}
\mathcal{L}_{R} = \sum_{\{R, \bar{R}\}}\Big(\min(R, 1) - \min(\bar{R}, 1)\Big)^2,
\label{eq:renderinLoss}
\end{equation}
where $\bar{R}$ is glossy sphere rendered with a ground truth HDR environment map. The reason for clamping is that in practice, most devices only have LDR displays. Therefore, over-penalizing the HDR pixels may not improve AR applications but can cause unstable training. Note that regardless of the clamping, the lighting we recover is HDR. The final loss function is a linear combination of the 2 loss functions. 
\begin{equation}
\mathcal{L}_{\text{single} } =  \mathcal{L}_{L_2} + \epsilon_{R} \mathcal{L}_{R},
\end{equation}
where $\epsilon_{R}$ is set to be 0.3 in our experiments. Note that we do not use adversarial loss to hallucinate the details of the unseen parts of the scene. We find that the hallucinated details may not create more realistic reflection in virtual object insertion and may cause temporally inconsistent lighting prediction, as shown in experiments.

\begin{figure}
\centering
\animategraphics[width=\columnwidth,poster=first]{2}{figures/depth/frame}{00}{10}
\caption{Comparison of blending models with and without depth panorama input $\{L_{D}^{i}\}$. Without depth input, the blending network cannot infer that the nearby white table and sofa is closer to the mirror sphere and therefore will occlude the surfaces further away seen in the later frames. Please click the figure to see the animation. }
\label{fig:depth}
\end{figure}

\begin{figure*}
\centering
\animategraphics[width=\textwidth,poster=first]{2}{figures/RNN/frame}{00}{12}
\caption{Comparison between our single-view lighting prediction from each individual frame and our RNN-based video lighting prediction. We observe that our RNN-based framework can effectively accumulate observation from multiple frames to predict HDR environment maps with more detailed reflection and more high-frequency lighting, while keeping the transition smooth. Note that even though the input depth maps are not fully temporal consistent, as shown in the detailed reflection in our single-view prediction, our framework still generates temporally consistent lighting prediction. Please click the figure to see the animation. }
\label{fig:rnn}
\end{figure*}

\subsection{Temporally consistent indoor lighting prediction}
We now extend our network to utilize video inputs for improved lighting prediction while maintaining temporal consistency. More specifically, our lighting prediction must be stable to avoid obvious flickering artifacts across frames, but as the camera motion accumulates, our network should gradually improve the lighting prediction quality based on new input frames. Intuitively, lighting prediction may be improved using new input frames in two aspects. First, as new inputs cover a larger region, the network can recover more details for this region, which can help render more realistic reflections on mirror or highly glossy surfaces. Second, even for a region that has never been seen, more inputs may still allow the network to reason about the invisible part of the scene using global illumination effects, including shadows, highlights, and color bleeding, which may benefit the reconstruction of important invisible light sources. 

\paragraph{3D RNN for temporally consistent SGLV reconstruction} A na\"{i}ve method to handle the video sequence would be to fuse the initial volumes and send them to the 3D encoder-decoder network. However, this method offers no explicit smoothness constraint and may suffer from inconsistent depth prediction across different views, causing blurry or flickering lighting prediction. Hence, we train RNNs to update our lighting prediction. Since our 3D encoder-decoder has only two skip connections, we only use two 3D gated recurrent unit (GRU) networks \cite{cho2014learning} $\mathbf{SGGRU}_0$ and $\mathbf{SGGRU}_1$. Similar network architectures are used to update 3D feature volumes in \cite{choy20163D2N2} for 3D reconstruction and \cite{sitzmann2019deepvoxels} for view synthesis. Let $V^i$ represent the volumes for frame $i$, we will have
\begin{eqnarray}
V_{f_0}^{i}, V_{f_1}^{i} &=& \mathbf{SGEncoder}(\tilde{V}^{i}_{\mathbf{c}}, \tilde{V}^{i}_{\alpha}, \tilde{V}^{i}_{e} ) \label{eq:3dRNN}\\
V_{h_0}^{i} &=& \mathbf{SGGRU}_0(V_{h_0}^{i-1}, V_{f_0}^{i}) \\
V_{h_1}^{i} &=& \mathbf{SGGRU}_1(V_{h_1}^{i-1}, V_{f_1}^{i})
\end{eqnarray}
where $V_{h_0}$ and $V_{h_1}$ are hidden states for $\mathbf{SGGRU}_0$ and $\mathbf{SGGRU}_1$, which are sent to the decoder. 
\begin{equation}
V_{\mathbf{c}}^i, V_{\alpha}^i, V_{\mathbf{w}}^i, V_{\lambda}^i, V_{\hat{\mathbf{s}}}^i, V_{u}^i = \mathbf{SGDecoder}(V_{h_0}^{i}, V_{h_1}^{i}) \label{eq:3dMerge}
\end{equation}
where $V_u$ is a single channel update volume to merge the prediction from the current frame with the prediction from the prior frame. Our final SGLV prediction for frame $i$ is computed as
\begin{eqnarray}
V_{\mathbf{x}}^{i} &=& V_{\mathbf{x}}^i (1 - V_{u}^i) + V_{u}^i V_{\mathbf{x}}^{i-1} \\
\mathbf{x} &\in& \{\mathbf{c}, \alpha, \mathbf{w}, \mathbf{\hat{s}}, \lambda\},
\end{eqnarray}
For the first frame with $i=0$, we directly use the 3D encoder-decoder trained for single image lighting prediction to reconstruct SGLV and the intermediate feature volumes $V_{f_0}^{0}$ and $V_{f_1}^{0}$. 

\paragraph{2D RNN for depth-aware blending} Similarly, to train the blending network to handle video inputs, we use two 2D GRU networks $\mathbf{BlendGRU}_0$ and $\mathbf{BlendGRU}_1$ to update the intermediate feature maps $F^{i}_{0}$ and $F^{i}_{1}$. However, in contrast to our single view blending network, our video blending network has to be depth-ware to reason about occlusion. Figure \ref{fig:depth} shows an example where the white table and the right sofa are close to the inserted mirror sphere but are not seen in later frames. The blending module, therefore, needs to "memorize" the occlusion without updating the detailed reflection from surfaces that are further away, which is only possible when provided with depth information. 

To make our 2D RNN blending network depth-aware, we further render a partial depth panorama $\tilde{L}_{D}^{i}$ using the partial mesh created from the depth map $D^{i}$. In addition, we introduce $\hat{L}_{D}^{i-1}$, which we define as the accumulated partial depth panorama from frames 0 to $i-1$ as shown in \eqref{eq:depthAccu}. By comparing the two partial depth panoramas $\tilde{L}_{D}^{i}$ and $\hat{L}_{D}^{i-1}$, we hope the blending network can learn to figure out the occlusion order of different regions. The encoder of our video blending network is defined as
\begin{equation}
F_0^i, F_1^i = \mathbf{BlendEncoder}(L^{i-1}, \dot{L}^{i}, \tilde{L}^{i}, \tilde{L}_{M}^{i}, 1 - \tilde{L}_{M}^{i}, \tilde{L}_{D}^{i}, \hat{L}_{D}^{i-1} )
\end{equation}
where $L^{i-1}$ is the blended final HDR environment map predicted from the last frame and $\tilde{L}_{M}^{i}$ is the binary mask indicating visible regions for frame $i$. The output feature maps $F_0$ and $F_1$ are then updated by two 2D GRU networks, as shown below.
\begin{eqnarray*}
H_{0}^{i} &=& \mathbf{BlendGRU}_0(H_{0}^{i-1}, F_{0}^{i}) \\
H_{1}^{i} &=& \mathbf{BlendGRU}_1(H_{1}^{i-1}, F_{1}^{i}) 
\end{eqnarray*}
The updated $H_0$ and $H_1$ are sent to the decoder to predict the blending weight $L_{M}^{i}$, 
\begin{equation}
L^{i}_{M} = \mathbf{BlendDecoder}(H_{0}^{i}, H_{1}^{i} ),
\end{equation}
which is used to blend the detailed partial environment map $\tilde{L}^{i}$ into our final HDR environment map prediction $L^{i}$ as shown in \eqref{eq:videoBlend}. To make our blending model robust to slight depth flickering, we adopt a conservative strategy by keeping the observed detailed reflection unchanged unless the newly visible surfaces is significantly closer to the camera, causing new occlusion. 
\begin{equation}
L^{i}_{M} = \max(L^{i}_{M} - \mathbf{1}(\hat{L}_{D}^{i-1} - \tilde{L}_{D}^{i} < 0.25), 0) 
\end{equation}
where $\mathbf{1}(\cdot)$ is an indicating function. 

Further, since the blending network is lightweight, we only use it to update the visible regions of the environment map, which is a relatively easier task. For the invisible regions, we directly use the $\dot{L}^{i}$ rendered from SGLV as the final prediction. To memorize the regions that have been seen in the past frames, we introduce the accumulated blending weight $\hat{L}_{M}^{i-1}$, which is the summation of blending weight from frame 0 to $i-1$ as shown in \eqref{eq:maskAccu}. Our environment map updating process can be written as
\begin{equation}
L^{i} = L^{i}_{M} \tilde{L}^{i} + (1 - L^{i}_{M})\left((1 - \hat{L}_{M}^{i-1}) \dot{L}^{i} + \hat{L}_{M}^{i-1}L^{i-1}  \right) \label{eq:videoBlend}
\end{equation}
Finally, we update the accumulated partial depth panorama $\hat{L}_{D}^{i}$ and the accumulated blending weight $\hat{L}_{M}^{i}$. 
\begin{eqnarray}
\hat{L}^{i}_{D} &=& L_{M}^{i} \tilde{L}^{i}_{D} + (1 - L_{M}^{i}) \hat{L}_{D}^{i-1} \label{eq:depthAccu}\\
\hat{L}^{i}_{M} &=& \min(\hat{L}^{i-1}_{M} + L_{M}^{i}, 1) \label{eq:maskAccu}
\end{eqnarray}

An alternative way to accumulate details of visible surfaces is running multi-view stereo methods \cite{schoenberger2016mvs} to reconstruct a mesh from a video sequence. However, this process may take several minutes to reconstruct the mesh while our current simple blending method can run in real-time, as shown in Table \ref{tab:time}, which is more suitable for building a mobile AR application with limited computational resources.

\paragraph{Loss functions.} To train our network for handling video inputs, we keep the per-pixel lighting loss $\mathcal{L}_{L_2}$ and the rendering loss $\mathcal{L}_{R}$ unchanged but add a new smoothness loss $\mathcal{L}_{sm}$ to enhance the temporal consistency. We define the smoothness loss to be the $\log$ $L_2$ distance between the predicted HDR environment maps from consecutive frames. 
\begin{equation*}
\mathcal{L}_{sm} = \sum_{\{L^{i}, L^{i-1}\}}\Big(\log(L^i + 1) - \log(L^{i-1} + 1) \Big)^2.
\end{equation*}
The final loss function is 
\begin{equation*}
\mathcal{L}_{video} =  \mathcal{L}_{L_2} + \epsilon_{R} \mathcal{L}_{R} + \epsilon_{sm} \mathcal{L}_{sm},
\end{equation*}
where $\epsilon_{R} = 0.3$ as before and $\epsilon_{sm}$ is set to be $0.01$.

Figure \ref{fig:rnn} compares our single view lighting prediction framework and our RNN-based video lighting prediction framework. For both frameworks in this example, we use depth maps from ARKit as inputs. We notice that our RNN-based framework can effectively accumulate the newly observed details from later frames while maintaining temporal consistency. Even though the invisible light sources are never seen in the whole video sequence, our framework manages to predict sharper HDR light sources while keeping them at almost the same location, leading to steadily sharper and more realistic shadows. On the contrary, the single view framework predicts inconsistent lighting across different frames, causing the reflection and shadows of the inserted mirror sphere to jump across frames. In addition, we may see that the detailed reflection of our single view lighting prediction twists noticeably as we move our camera, which is caused by the flickering of ARKit depth prediction. On the contrary, our depth-aware RNN-based blending model is robust to such flickering and can create stable detailed reflection across the whole video sequence.

\subsection{Dataset creation and implementation details}
\label{sec:details}

\paragraph{Dataset creation.} To train our framework, we augment the OpenRooms dataset \cite{li2020openrooms}, which is a large-scale, high-quality synthetic indoor dataset, created with realistic materials and lighting, and rendered using an OptiX-based path tracer. The original OpenRooms dataset already contains spatially varying, per-pixel HDR environment maps densely sampled on the scene surface. However, those HDR environment maps are of a too low resolution ($16\times 32$) and none of them locates in the middle of the room. Therefore, we randomly sample 3 positions inside each camera frustum of 120K images and render an HDR environment with a resolution of $120\times 240$, leading to 360K much higher resolution environment maps in total. 

To create video sequences, we follow the two-body method used in \cite{mccormac2017scenenet, li2018interiornet} to simulate smooth camera trajectories. We render $37,680$ video sequences in total, each of which contains 31 frames with a resolution of $240\times 320$. The average camera distance and the rotation angle between consecutive frames are $0.09$m and $4.78^{\circ}$, respectively. For each video sequence, we randomly sample 3 positions in the camera frustum of the first frame and render HDR environment maps at these positions as the ground truth supervision. 

\begin{table}[t]
\centering
\begin{tabular}{|c|c|c|}
\hline
     & Range & Resolution \\
\hline
x & [-1.1$D_{\max}$, 1.1$D_{\max}$] & 84 \\
\hline
y & [-0.8$D_{\max}$, 0.8$D_{\max}$] & 60  \\
\hline
z & [-1.2$D_{\max}$, 0.5$D_{\max}$] & 64 \\
\hline
\end{tabular}
\caption{Volume range and resolution in $x$, $y$ and $z$ dimension. $D_{\max}$ is the maximum depth value in the first frame.}
\label{tab:volumeSize}
\end{table}

\begin{table}[t]
\centering
\begin{tabular}{|c|c|c|c|}
\hline
& SGLV & Blending & Joint \\
\hline
Single View  & 7 & 2 & 2\\
\hline
Video & 3 & 3 & 2 \\  
\hline
\end{tabular}
\caption{The number of training epochs for each step.}
\label{tab:training}
\end{table}

\begin{table}[t]
\centering
\begin{tabular}{|c|c|c|c|c|c|}
\hline
Volume & SGLV  & Volume & Blend- & 3D & 2D \\
initialization & prediction & rendering & ing  & RNN & RNN \\
\hline
4.1 ms  & 25 ms & 22 ms & 5.2 ms & 19 ms & 2.1 ms \\
\hline
\end{tabular}
\caption{Time consumption of every step of our framework. The total time to process one frame is around 77 ms.}
\label{tab:time}
\end{table}

\paragraph{Volume configuration.} We decide the location and orientation of our volume based on the camera position, with the camera center being the original point, and the right, up, backward directions of the image plane being the $x$, $y$, $z$ axes. We decide the size and resolution of our volume based on the maximum depth value $D_{\max}$ in the depth map $D$, as summarized in Table \ref{tab:volumeSize}. We empirically find that this configuration allows us to cover most light sources in our volume. When the input is a video sequence, we build the volume based on the first frame and fix it through the entire video sequence. Our total number of voxels is only 25\% of Lighthouse \cite{srinivasan2020lighthouse} and 15\% of Wang et al. \cite{wang2021learning}, which makes our model much more memory efficient.

\paragraph{Training details} We use the Adam optimizer \cite{adam} with a learning rate of $10^{-4}$ and batch size 1 to train all our networks. We first train the networks for single view indoor lighting prediction and use the trained models as the initialization for handling video inputs. We find that this strategy allows us to significantly reduce the total training time. For both single view and video cases, we train the 3D encoder-decoder for SGLV reconstruction first, then fix it to train the blending network, and finally, jointly fine-tune the two networks together. The number of iterations for each step is summarized in Table \ref{tab:training}. 

To test the model performances with imperfect depth inputs, we train two versions of our lighting prediction models, one with the ground truth depth maps and the other with predicted depth maps. For single view lighting prediction, we choose 3D-ken-burn by Niklaus et al. \cite{niklaus20193d} for depth prediction because of its good performances on indoor scene datasets \cite{silberman2012indoor}. For video lighting prediction, we choose Neural RGBD by Liu et al. \cite{liu2019neural} for depth prediction. We fine-tune both models on our synthetic dataset. Some more recent works demonstrate that the transformer architecture can achieve more accurate depth prediction \cite{luo2020consistent, zhang2021consistent}. Other works propose to predict temporally consistent depth maps by fine-tuning a pre-trained network for every video sequence. While these new methods may improve lighting prediction quality, it is not our focus to comprehensively explore all different depth prediction methods.

\paragraph{Real data} For all real experiments on single image lighting prediction, we use depth prediction from Niklaus et al. \cite{niklaus20193d} to demonstrate that we can handle any legacy photos. For video lighting prediction, we use camera poses and depth maps directly obtained from iPad ARkit. While the depth maps from ARKit may not be completely temporally consistent, we observe that our method is robust to such imperfect depth inputs. We sample 30 frames from around 10 seconds video sequences as the input to our network. Once we get the lighting prediction of the 30 frames, we use simple bilinear interpolation to generate the lighting prediction for every frame in the whole video sequence and render our object insertion results.

\paragraph{Time consumption} We test the time consumption of our framework on an Nvidia 2080 Ti GPU. The results are summarized in Table \ref{tab:time}. We observe that the majority of time is spent on reconstructing SGLV and using volume ray tracing to render HDR environment maps. The total time consumption for one frame is around 77 ms, which means our method can run at 13 fps.  

%% file: sources/experiment.tex
\section{Experiments}
\label{sec:experiment}

We first present our results of spatially consistent single view lighting prediction and then our results of spatiotemporally consistent video lighting prediction, on both synthetic and real data. 

\begin{figure}[t]
\centering
\includegraphics[width=\columnwidth]{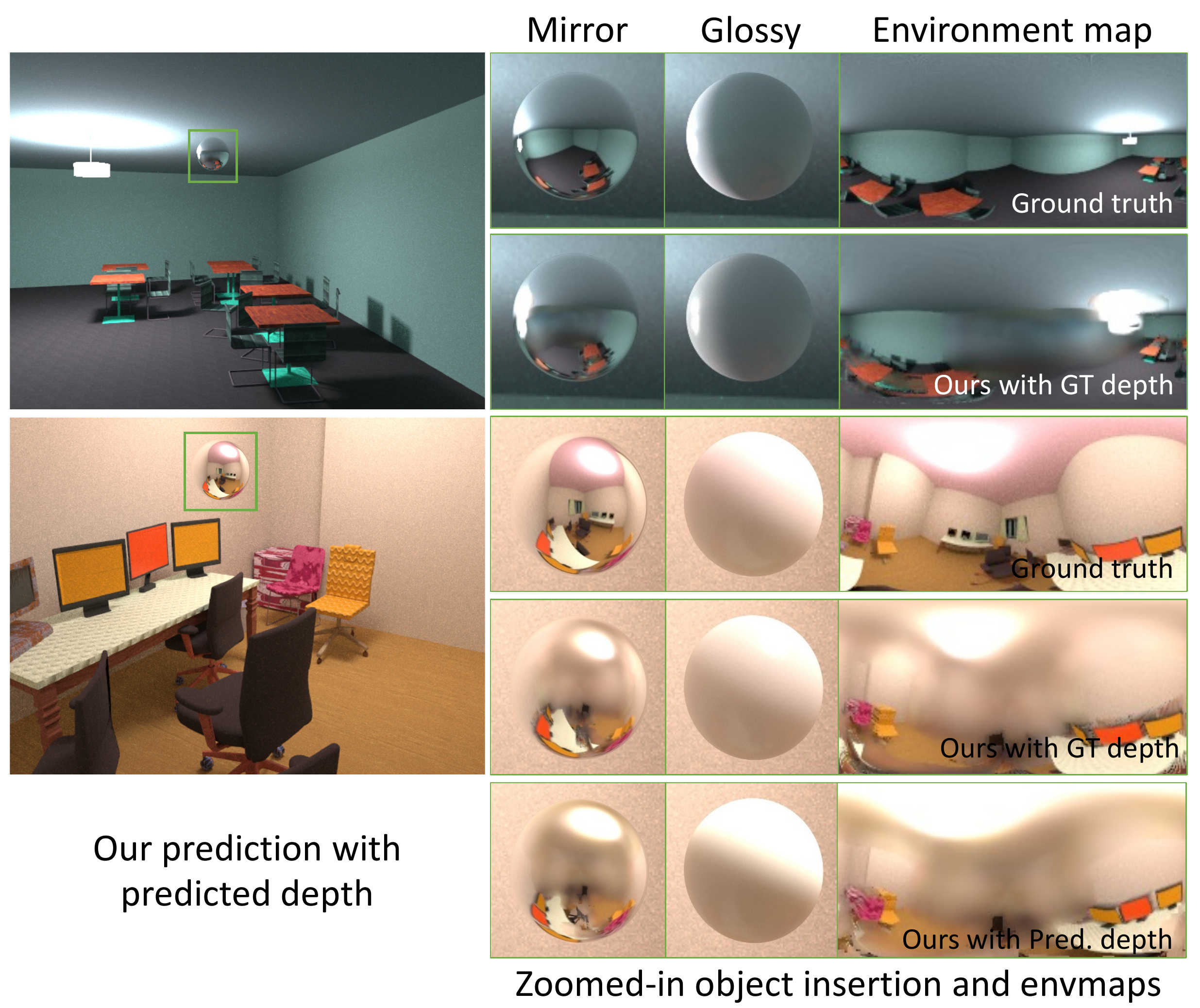}
\caption{Single view lighting prediction on our synthetic dataset. Our method generates high-quality lighting prediction with detailed reflection as well as HDR visible and invisible light sources, which can support realistic rendering of both glossy and mirror materials. Our lighting predictions using predicted depth maps is comparable to those using ground truth depth maps, with only slight distortion of the reflection and blur of the HDR light sources.}
\label{fig:singleSyn}
\end{figure}

\begin{figure*}[t]
\centering
\includegraphics[width=\textwidth]{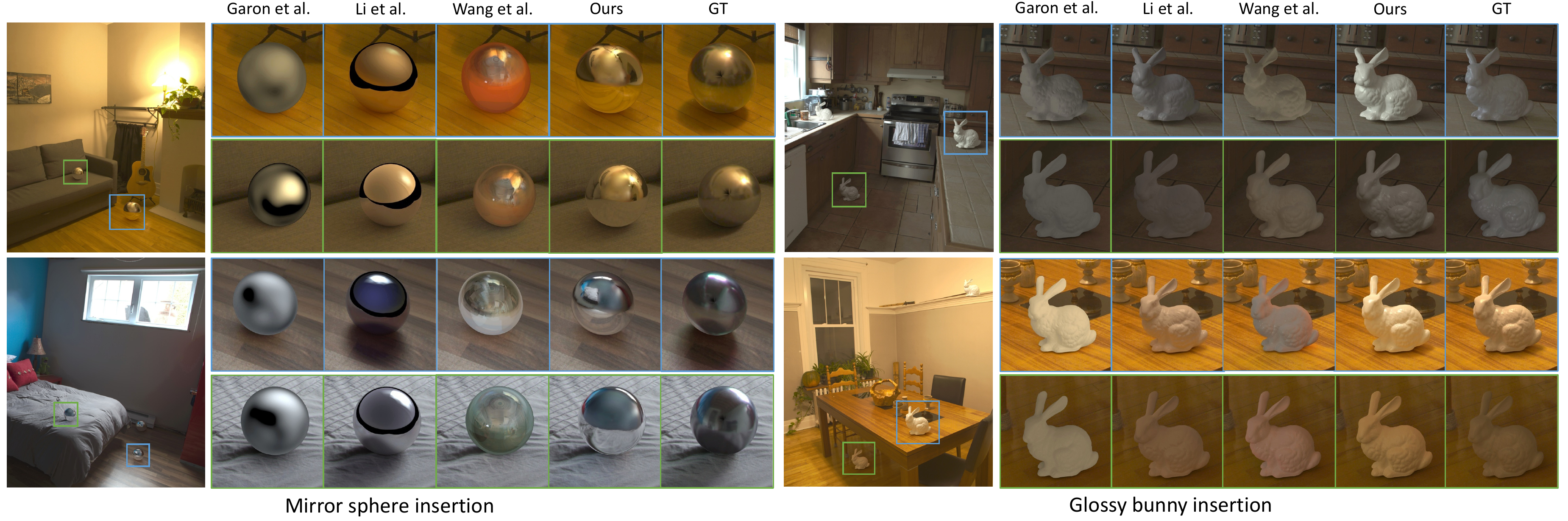}
\caption{Comparison of single view lighting prediction for object insertion on Garon et al. dataset \cite{garon2019fast}. Our framework can recover detailed reflection as well as visible and invisible HDR light sources, with various light transport effects such as directional lighting, global illumination, and occlusion being correctly modeled, which leads to realistic rendering for both mirror and glossy objects at any given location in the scene. }
\label{fig:singleInsertion}
\end{figure*}

\begin{table}[t]
\centering
\begin{tabular}{|l|c|c|c|c|}
\hline
\multirow{2}{*}{} & \multicolumn{2}{|c|}{SGLV}  & RGB$\alpha$ & Pred.   \\ 
\cline{2-4}  
& Init. $\dot{L}$  & Blended $L$  & Blended $L$  & depth\\
\hline
Env $\mathcal{L}_{L_2}$ ($10^{-1}$)  & 9.52 & \textbf{9.02} & 9.05 & 10.4\\
\hline
Render $\mathcal{L}_{R}$ ($10^{-2}$) & 2.92 & \textbf{2.78} & 2.88 & 3.23\\
\hline
\end{tabular}
\caption{Quantitative results of single view lighting prediction on our synthetic dataset. We observe that our hybrid framework can greatly improve accuracy compared to the pure volume-based framework. Our SGLV representation can better model HDR light sources compared to RGB$\alpha$ volume, which leads to smaller rendering loss. We include our lighting prediction accuracy with predicted depth in the last column.}
\label{tab:synSingle}
\end{table}

\begin{table}[t]
\centering
\begin{tabular}{|c|c|c|}
\hline
& Glossy bunnies & Mirror spheres \\
\hline
\cite{gardner2017indoor} & 61.7\% & 78.2\% \\
\hline 
 \cite{garon2019fast} & 59.9\%  & 77.6\%  \\
\hline
\cite{li2020inverse} & 46.3\%  & 77.1\%  \\
\hline
\cite{wang2021learning} & 66.6\% & 66.6\% \\
\hline
\end{tabular}
\caption{User study of object insertion on the Garon et al. \cite{garon2019fast} dataset. For all 20 indoor scenes in the dataset, we place mirror spheres and glossy bunnies at multiple locations inside the scenes and render them with predicted HDR environment maps. Given randomly selected pairs of object insertion results, users will be asked to judge which result is more realistic. Each user will evaluate 20 pairs and 200 users recruited through Amazon Turk joined this study. We report the percentage of users who consider our method to be more realistic. For mirror sphere insertion, our results are much better compared to prior state-of-the-arts, as 60\% to 80\% users agree that they are more realistic. For glossy bunny insertion, our results are only slightly worse than Li et al. \cite{li2020inverse} but better than other works. }
\label{tab:userStudy}
\end{table}

\subsection{Singe image indoor lighting prediction}

\paragraph{Synthetic data} Quantitative and qualitative results of single view lighting prediction on our synthetic dataset are summarized in Figure \ref{fig:singleSyn} and Table \ref{tab:synSingle}. From Figure \ref{fig:singleSyn}, we observe that our method can recover the details of the reflection from visible surfaces as well as the HDR light sources, either visible or invisible. As for the invisible reflection of the scene, our method cannot recover the details but can infer correct color distribution and also maintain a smooth, natural transition between visible and invisible regions. Using our lighting prediction, we can render both mirror and glossy spheres into the scene with appearance closely matching the ground truths.  From Table \ref{tab:synSingle}, we observe that our hybrid framework with the blending network can produce much more accurate lighting prediction, as also shown in the real examples in Figure \ref{fig:blending}. In addition, we observe that our SGLV representation has lower rendering loss of glossy sphere compared to vanilla RGB$\alpha$ volume. This implies that our representation can better handle HDR lighting and directional lighting, which is also shown in the real example in Figure \ref{fig:SGvsRGB}. 

\begin{figure}
\centering
\animategraphics[width=\columnwidth,poster=first]{2}{figures/spCons/frame}{00}{16}
\caption{We demonstrate the spatial consistency of our lighting prediction by placing moving glossy and mirror spheres into two real scenes. Please click the figure to see the animation.}
\label{fig:spCons}
\end{figure}

\paragraph{Real data}

We compare our method with prior works on Garon et al. dataset \cite{garon2019fast}, which is widely used for evaluating indoor lighting. The dataset contains 20 images with lighting selected at different locations in the scenes. We use HDR environment maps predicted by different methods to render virtual mirror spheres and glossy bunnies into the scenes, to compare the level of photorealism of object insertion under different materials.  Qualitative results are summarized in Figure \ref{fig:singleInsertion}. We see that our method can recover both detailed reflection and HDR light sources accurately, generating realistic object insertion for both mirror and glossy materials, with high-quality reflection, specular highlights, and shadows. It models challenging visual effects effectively, such as global illumination, occlusion, and directional lighting. For example, in the bottom-right scene, the bunny under the table is much darker compared to the bunny on the table due to the occlusion of light sources; in the top-right scene, the bunny on the platform has realistic highlights and shadows caused by sunlight coming through the window. In contrast, a recent method \cite{wang2021learning} that uses a similar spherical Gaussian volume as its lighting representation fails to reconstruct realistic nearby reflection even though they use around 6.5 times more voxels compared to our hybrid representation, as shown in the mirror sphere insertion results. They also cannot recover HDR lighting as well as our method, probably because their dataset lacks the direct supervision from ground truth HDR environment maps. This causes their glossy object insertion results to be darker than ground truth, with blurry specular highlights and shadows. Prior state-of-the-art \cite{li2020inverse} cannot recover the detailed reflection, leading to poor-quality rendering for mirror spheres. It also cannot guarantee spatial consistency. Earlier methods cannot recover high-frequency HDR lighting \cite{garon2019fast} or spatially-varying lighting \cite{gardner2017indoor}.

To evaluate the object insertion quality quantitatively, we conduct a user study on Amazon Turk. For mirror sphere insertion and glossy bunny insertion, we hire 200 Amazon Turkers respectively, present them with 20 pairs of object insertion results, with each pair generated by two randomly selected methods, and ask them which one looks more realistic. We report the percentage of people thinking our results are more realistic in Table \ref{tab:userStudy}. The quantitative numbers match our observation in Figure \ref{fig:singleInsertion}, which shows that our method outperforms all prior state-of-the-art in mirror sphere insertion, including the most recent volume-based lighting estimation method \cite{wang2021learning}, and is only slightly worse compared to \cite{li2020inverse} in glossy bunny insertion. However, our lighting estimation framework is more general compared to \cite{li2020inverse}, which can only predict environment map at the surface with no guarantee of spatial consistency while we can predict spatially-consistent lighting at arbitrary locations. 

In Figure \ref{fig:spCons}, we demonstrate the spatial consistency of our lighting estimation by moving virtual glossy and mirror spheres inside the scene. The animation shows that the reflection, specular highlights and shadows of the spheres are all consistent as we change their locations.

\begin{figure}
\centering
\animategraphics[width=\columnwidth,poster=first]{2}{figures/videoSyn/frame}{00}{10}
\caption{Video lighting prediction on our synthetic dataset. Our method can progressively add visible details while making the predicted light source temporally consistent. Our rendered glossy and mirror spheres are close to the ground truths. In contrast, Lighthouse \cite{srinivasan2020lighthouse} cannot recover invisible light sources or achieve temporal consistency. Even with predicted depth, our method can still produce higher quality, temporally consistent lighting prediction with only slight blur and distortion. Please click the figure to see the animation. }
\label{fig:videoSyn}
\end{figure}

\begin{figure*}
\centering
\animategraphics[width=\textwidth,poster=first]{2}{figures/videoReal/frame}{00}{10}
\caption{Comparisons of video lighting prediction and object insertion on real data captured by an iPad. Our method significantly outperforms prior state-of-the-art \cite{srinivasan2020lighthouse} to predict more temporally consistent lighting, more accurate HDR light sources and reflection, which allows more realistic shadows and specular highlight in object insertion with different materials. Please click the figure to see the animation. }
\label{fig:videoReal}
\end{figure*}

\begin{figure*}
\centering
\animategraphics[width=\textwidth,poster=first]{2}{figures/videoReal2/frame}{00}{10}
\caption{Another video lighting prediction example. Our method memorizes the visible lamp that only appears in the first few frames, leading to consistent specular highlights and shadows across the whole video sequence. Please click the figure to see the animation. }
\label{fig:videoReal2}
\end{figure*}

\begin{table}[t]
\centering
\renewcommand{\tabcolsep}{0.12cm}
\small
\begin{tabular}{|c|c|c|c|c|c|c|}
\hline
\multicolumn{2}{|c|}{Frame interval}  & [1,8) & [8,14) & [14,20) & [20,26) & [26,31] \\
\hline
\multirow{4}{*}{HDR env} & Gt. depth & 0.63 & 0.61 & 0.60 & 0.59 & \textbf{0.59} \\ 
\cline{2-7}
 & Pred. depth & 0.85 & 0.80 & 0.78 & 0.77 & \textbf{0.77} \\
\cline{2-7}
& No RNN & \textbf{0.93} & 1.15 & 1.14 & 1.15 & 1.19\\ 
\cline{2-7} 
& Simple Avg. & \textbf{0.71} & 0.75 & 0.79 & 0.81 & 0.84 \\
\cline{2-7}
$\log$ $L_2$ ($10^{-1}$) & Lighthouse & \textbf{1.39} & 1.42 & 1.42 & 1.41 & 1.42 \\ 
\hline
\multicolumn{2}{|c|}{Frame interval}  & [1,8) & [8,14) & [14,20) & [20,26) & [26,31] \\
\hline
\multirow{4}{*}{LDR env} & Gt. depth & 2.5 & 2.4 & 2.3 & 2.3 & \textbf{2.2} \\ 
\cline{2-7}
 & Pred. depth & 3.8 & 3.5 & 3.4 & 3.3 & \textbf{3.3} \\
\cline{2-7}
& No RNN & \textbf{4.4} & 6.1 & 6.0 & 6.2 & 6.6 \\ 
\cline{2-7} 
& Simple Avg. & \textbf{3.2}  & 3.5 & 3.8 & 4.0 & 4.2 \\
\cline{2-7}
 $L_2$ ($10^{-2}$) & Lighthouse & \textbf{6.7} & 6.9 & 7.0 & 7.0 & 7.0 \\ 
\hline
\multicolumn{2}{|c|}{Frame interval}  & [1,8) & [8,14) & [14,20) & [20,26) & [26,30] \\
\hline
\multirow{4}{*}{Smoothness} & Gt. depth & 1.38 & 0.74 & 0.58 & 0.52 & \textbf{0.49} \\ 
\cline{2-7}
 & Pred. depth & 2.12 & 0.86 & 0.68 & 0.61 & \textbf{0.56} \\
\cline{2-7}
& No RNN & 12.1 & 6.3 & 6.1 & 6.0 & \textbf{6.0} \\ 
\cline{2-7} 
& Simple Avg. & 0.76 & 0.41 & 0.35 & 0.34 & \textbf{0.33} \\
\cline{2-7}
 $\log$ $L_2$ ($10^{-3}$) & Lighthouse & 3.85 & 2.38 & 2.27 & 2.19 & \textbf{2.16} \\ 
\hline
\end{tabular}
\caption{Quantitative comparisons of video lighting prediction on our synthetic dataset. We divide our 31-frame video sequences equally into five intervals and report the errors of each interval. The interval with the lowest error is marked with bold font. All three metrics show that with both ground truth and predicted depth maps as inputs, our method can consistently improve lighting prediction accuracy as it sees larger parts of scenes. On the contrary, prior state-of-the-art \cite{srinivasan2020lighthouse} predicts less accurate LDR environment maps in the first few frames and cannot improve the prediction accuracy with more inputs. }
\label{tab:synVideo}
\end{table}

\subsection{Temporally consistent indoor lighting prediction}

\paragraph{Synthetic data} Figure \ref{fig:videoSyn} shows qualitative results of video lighting prediction on our synthetic dataset. Our method manages to produce high-quality HDR environment map prediction, which enables realistic rendering of mirror and glossy spheres similar to the ground truths. In comparison to single view lighting prediction, our video lighting prediction includes much more details as our input frames cover a larger region of the room. As shown by the animation in Figure \ref{fig:videoSyn}, our method manages to maintain reasonable temporal consistency as it refines the lighting prediction with new inputs. In the first example, our method memorizes both the invisible lamp on the top and the window which cannot be seen in the last few frames, through the whole video. In the second example, our method keeps the location of the invisible lamp unchanged and makes it sharper as it looks around the room. We also show our video lighting prediction with predicted depth in the second example, which presents more blurry HDR light sources and slightly distorted details but still has smooth inter-frame transition even though the depth inputs from different frames may not be accurate or consistent. On the contrary, prior state-of-the-art methods' lighting prediction changes drastically as we move the camera. Besides, it cannot recover invisible light sources, which is very important for lighting estimation. 

Our quantitative comparisons are summarized in Table \ref{tab:synVideo}. We compare our method with prior volume-based lighting prediction method Lighthouse \cite{srinivasan2020lighthouse} and two baselines. In the first baseline "No RNN", we remove the 2D and 3D GRU network modules and predict lighting independently from each input frame. In the second baseline "Simple Avg.", we simply accumulate our independent lighting prediction by computing their average. We report three errors, the $\log$ $L_2$ HDR environment map error $\mathcal{L}_{L_2}$ and smoothness error $\mathcal{L}_{sm}$ are used to train our network. To make a fair comparison with prior work \cite{srinivasan2020lighthouse} that can only predict LDR environment maps, we also report LDR environment map loss in the second group where we clamp the lighting prediction to the range of 0 to 1 and then compute its $L_2$ error from the ground truth. The quantitative comparisons show that our method can consistently reduce the lighting prediction errors with new input frames, even with imperfect predicted depth maps. On the contrary, the prior volume-based method predicts less accurate lighting as the camera drifts away. In all three metrics, our method with ground truth or predicted depths outperforms prior work by a large margin. Two simple baselines cannot progressively improve lighting predictions with new input frames either, which further demonstrates the effectiveness of our RNN modules.

\paragraph{Real data} Figure \ref{fig:videoReal} and Figure \ref{fig:videoReal2} show three examples of our video lighting prediction on real data captured with an iPad. We sample 10 frames of prediction and include the full video sequence in the supplementary material. The depth maps and camera poses are obtained from ARKit. We observe that even though our method is trained on synthetic videos only, it generalizes well to real videos, as it can still progressively improve the lighting prediction quality by adding more details to the reflection from visible surfaces as well as refining the HDR light sources and reflection from invisible surfaces. 

More specifically, in the first example in \ref{fig:videoReal}, our method refines the lighting prediction as a light source that is invisible in the first few frames and, therefore, not correctly predicted, gradually emerges in later frames. Our method successfully adds the new light sources as it shows up. Meanwhile, the other invisible light source, which has been predicted correctly in the beginning, remains stable across the video sequence. In the third example, our method predicts two invisible light sources (a lamp on the top and a window on the left) accurately from the first few frames and keeps refining them as more inputs are available. Through this process, the shadows and highlights on both the mirror spheres and the "knight" are reasonably stable. In Figure \ref{fig:videoReal2}, our method "memorizes" the visible light source that only appears in the first half of the video. As a result, the shadows and highlights are consistent across the whole video sequence. Note that even though the depth maps from ARKit may flicker slightly, as shown in Figure \ref{fig:rnn}, our blending model is robust to such small errors and generates stable details across all three video sequences. 

In contrast, prior work Lighthouse \cite{srinivasan2020lighthouse} cannot predict temporally consistent lighting. More importantly, it cannot utilize later frames to refine its predictions. Therefore, as the camera drifts away, their predictions become worse. In addition, even for the first few frames, we observe that Lighthouse can only recover the reflection of visible surfaces. For invisible surfaces, it cannot predict either light sources or a correct color distribution that allows a smooth transition between visible and invisible reflection. On the contrary, our lighting prediction is much more natural and realistic on real data, mainly because our dataset (physically-based rendered OpenRooms dataset), network architecture and loss functions (rendering loss) are all physically motivated, causing our learning-based framework to have better generalization ability.  

%% file: sources/conclusion.tex
\section{Conclusion}
\label{sec:conclusion}

\paragraph{Limitations and future work} Our current method requires camera poses and depth maps as inputs. For single-view lighting prediction, we can achieve state-of-the-art performances with depth prediction from \cite{niklaus20193d}. For video lighting prediction, our method works well with the camera poses and low-resolution depth maps captured by iPad Arkit. How to predict temporally consistent depth maps is still an open problem \cite{luo2020consistent,zhang2021consistent} and even depth maps captured by ARKit flicker noticeably. However, as shown in several real examples in Figure \ref{fig:teaser}, Figure \ref{fig:rnn} and Figure \ref{fig:videoReal}, our method is robust to such flickering. In addition, if the input video sequence is too long, the camera poses from ARKit will be less accurate and cause errors in our lighting prediction. As higher quality sensors become more and more accessible on mobile devices, we suggest that these limitations may be less influential in the future. 

In addition, our method may fail in the presence of highly glossy or mirror materials. In Figure \ref{fig:rnn}, we can see that the reflection of highlight on the tablet gets blurred in our HDR environment map prediction as the camera moves away. This is because as we change our viewpoints, this highlight is no longer visible in later frames (Frame 13, for example). This suggests that we need a holistic framework to jointly reason about the geometry and materials of indoor scenes to achieve more accurate lighting prediction. 

Besides, our method cannot hallucinate detailed reflection from invisible surfaces. We argue that this is a too challenging problem and humans may not be very sensitive to these details. Prior method \cite{srinivasan2020lighthouse} tries to solve this problem by adding adversarial loss. However, as shown in both synthetic and real examples in Figure \ref{fig:videoSyn} and Figure \ref{fig:videoReal}, they only add random textures to the environment map prediction without substantially improving its visual quality. We also tried adversarial loss but observe that it deteriorates our model's generalization ability by adding weird colors in some real cases. On the contrary, our current method generally can predict the overall color distribution of invisible reflection with smooth transition between visible and invisible regions, leading to more visually convincing lighting prediction. How to synthesize realistic consistent invisible details can be an interesting direction for future exploration.

Finally, our current method is designed for indoor scenes only and will not generalize well to outdoor scenes. While our lighting representation can model directional lighting, our training data contain only indoor scene images. Moreover, the scales of outdoor scenes are much larger than indoor scenes, which will require us to modify our volume parameterization to handle scene structures that are far away from the camera. In addition, our video lighting prediction method can only handle static scenes but cannot handle dynamic scenes with changing lighting or moving objects.

\paragraph{Summary} We present a physically-motivated deep learning-based framework for spatiotemporally consistent HDR lighting prediction at arbitrary locations of indoor scenes, which enables realistic object insertion with different materials. Our method is the first that can utilize video sequences to improve lighting prediction, while explicitly considering temporal consistency as we progressively refine the outputs. The key novelties of our framework include: (1) a hybrid framework that can recover both visible details and HDR light sources as well as guarantee spatial consistency and (2) an RNN-based framework that can handle video sequences. Experiments show that our framework is more general, as it can handle more diverse inputs for different applications, and with comparable performances to prior state-of-the-arts.

%% file: main.bbl

\begin{thebibliography}{37}


\ifx \showCODEN    \undefined \def \showCODEN     #1{\unskip}     \fi
\ifx \showDOI      \undefined \def \showDOI       #1{#1}\fi
\ifx \showISBNx    \undefined \def \showISBNx     #1{\unskip}     \fi
\ifx \showISBNxiii \undefined \def \showISBNxiii  #1{\unskip}     \fi
\ifx \showISSN     \undefined \def \showISSN      #1{\unskip}     \fi
\ifx \showLCCN     \undefined \def \showLCCN      #1{\unskip}     \fi
\ifx \shownote     \undefined \def \shownote      #1{#1}          \fi
\ifx \showarticletitle \undefined \def \showarticletitle #1{#1}   \fi
\ifx \showURL      \undefined \def \showURL       {\relax}        \fi
\providecommand\bibfield[2]{#2}
\providecommand\bibinfo[2]{#2}
\providecommand\natexlab[1]{#1}
\providecommand\showeprint[2][]{arXiv:#2}

\bibitem[Akimoto et~al\mbox{.}(2022)]%
        {akimoto2022diverse}
\bibfield{author}{\bibinfo{person}{Naofumi Akimoto}, \bibinfo{person}{Yuhi
  Matsuo}, {and} \bibinfo{person}{Yoshimitsu Aoki}.}
  \bibinfo{year}{2022}\natexlab{}.
\newblock \showarticletitle{Diverse Plausible 360-Degree Image Outpainting for
  Efficient 3DCG Background Creation}. In \bibinfo{booktitle}{\emph{Proceedings
  of the IEEE/CVF Conference on Computer Vision and Pattern Recognition}}.
  \bibinfo{pages}{11441--11450}.
\newblock


\bibitem[Barron and Malik(2013)]%
        {barron2013intrinsic}
\bibfield{author}{\bibinfo{person}{Jonathan~T Barron} {and}
  \bibinfo{person}{Jitendra Malik}.} \bibinfo{year}{2013}\natexlab{}.
\newblock \showarticletitle{Intrinsic scene properties from a single {RGB-D}
  image}. In \bibinfo{booktitle}{\emph{Proc.~CVPR}}.
\newblock


\bibitem[Cho et~al\mbox{.}(2014)]%
        {cho2014learning}
\bibfield{author}{\bibinfo{person}{Kyunghyun Cho}, \bibinfo{person}{Bart
  Van~Merri{\"e}nboer}, \bibinfo{person}{Caglar Gulcehre},
  \bibinfo{person}{Dzmitry Bahdanau}, \bibinfo{person}{Fethi Bougares},
  \bibinfo{person}{Holger Schwenk}, {and} \bibinfo{person}{Yoshua Bengio}.}
  \bibinfo{year}{2014}\natexlab{}.
\newblock \showarticletitle{Learning phrase representations using RNN
  encoder-decoder for statistical machine translation}.
\newblock \bibinfo{journal}{\emph{arXiv preprint arXiv:1406.1078}}
  (\bibinfo{year}{2014}).
\newblock


\bibitem[Choy et~al\mbox{.}(2016)]%
        {choy20163D2N2}
\bibfield{author}{\bibinfo{person}{Christopher~B Choy}, \bibinfo{person}{Danfei
  Xu}, \bibinfo{person}{JunYoung Gwak}, \bibinfo{person}{Kevin Chen}, {and}
  \bibinfo{person}{Silvio Savarese}.} \bibinfo{year}{2016}\natexlab{}.
\newblock \showarticletitle{3d-r2n2: A unified approach for single and
  multi-view 3d object reconstruction}. In
  \bibinfo{booktitle}{\emph{Proc.~ECCV}}.
\newblock


\bibitem[Dastjerdi et~al\mbox{.}(2022)]%
        {dastjerdi2022guided}
\bibfield{author}{\bibinfo{person}{Mohammad Reza~Karimi Dastjerdi},
  \bibinfo{person}{Yannick Hold-Geoffroy}, \bibinfo{person}{Jonathan
  Eisenmann}, \bibinfo{person}{Siavash Khodadadeh}, {and}
  \bibinfo{person}{Jean-Fran{\c{c}}ois Lalonde}.}
  \bibinfo{year}{2022}\natexlab{}.
\newblock \showarticletitle{Guided Co-Modulated GAN for 360
  $\{$$\backslash$deg$\}$ Field of View Extrapolation}.
\newblock \bibinfo{journal}{\emph{arXiv preprint arXiv:2204.07286}}
  (\bibinfo{year}{2022}).
\newblock


\bibitem[Debevec(1998)]%
        {debevec1998rendering}
\bibfield{author}{\bibinfo{person}{Paul Debevec}.}
  \bibinfo{year}{1998}\natexlab{}.
\newblock \showarticletitle{Rendering synthetic objects into real scenes:
  Bridging traditional and image-based graphics with global illumination and
  high dynamic range photography}. In \bibinfo{booktitle}{\emph{Proc. SIGGRAPH
  98}}. \bibinfo{pages}{189--198}.
\newblock


\bibitem[Gardner et~al\mbox{.}(2019)]%
        {gardner2019parametric}
\bibfield{author}{\bibinfo{person}{Marc-Andr\'{e} Gardner},
  \bibinfo{person}{Yannick Hold-Geoffroy}, \bibinfo{person}{Kalyan Sunkavalli},
  \bibinfo{person}{Christian Gagn\'{e}}, {and}
  \bibinfo{person}{Jean-Fran\c{c}ois Lalonde}.}
  \bibinfo{year}{2019}\natexlab{}.
\newblock \showarticletitle{Deep Parametric Indoor Lighting Estimation}. In
  \bibinfo{booktitle}{\emph{Proc.~ICCV}}.
\newblock


\bibitem[Gardner et~al\mbox{.}(2017)]%
        {gardner2017indoor}
\bibfield{author}{\bibinfo{person}{Marc-Andr{\'e} Gardner},
  \bibinfo{person}{Kalyan Sunkavalli}, \bibinfo{person}{Ersin Yumer},
  \bibinfo{person}{Xiaohui Shen}, \bibinfo{person}{Emiliano Gambaretto},
  \bibinfo{person}{Christian Gagn{\'e}}, {and}
  \bibinfo{person}{Jean-Fran{\c{c}}ois Lalonde}.}
  \bibinfo{year}{2017}\natexlab{}.
\newblock \showarticletitle{Learning to predict indoor illumination from a
  single image}.
\newblock \bibinfo{journal}{\emph{ACM Trans. Graphics}} \bibinfo{volume}{9},
  \bibinfo{number}{4} (\bibinfo{year}{2017}).
\newblock


\bibitem[Garon et~al\mbox{.}(2019)]%
        {garon2019fast}
\bibfield{author}{\bibinfo{person}{Mathieu Garon}, \bibinfo{person}{Kalyan
  Sunkavalli}, \bibinfo{person}{Sunil Hadap}, \bibinfo{person}{Nathan Carr},
  {and} \bibinfo{person}{Jean-Fran{\c{c}}ois Lalonde}.}
  \bibinfo{year}{2019}\natexlab{}.
\newblock \showarticletitle{Fast Spatially-Varying Indoor Lighting Estimation}.
  In \bibinfo{booktitle}{\emph{Proc.~CVPR}}.
\newblock


\bibitem[Karis(2013)]%
        {brdfModel}
\bibfield{author}{\bibinfo{person}{Brian Karis}.}
  \bibinfo{year}{2013}\natexlab{}.
\newblock \showarticletitle{Real shading in Unreal Engine 4}.
\newblock \bibinfo{journal}{\emph{SIGGRAPH 2013 Courses: Physically Based
  Shading Theory Practice}} (\bibinfo{year}{2013}).
\newblock


\bibitem[Karsch et~al\mbox{.}(2014)]%
        {karsch-tog-14}
\bibfield{author}{\bibinfo{person}{Kevin Karsch}, \bibinfo{person}{Kalyan
  Sunkavalli}, \bibinfo{person}{Sunil Hadap}, \bibinfo{person}{Nathan Carr},
  \bibinfo{person}{Hailin Jin}, \bibinfo{person}{Rafael Fonte},
  \bibinfo{person}{Michael Sittig}, {and} \bibinfo{person}{David Forsyth}.}
  \bibinfo{year}{2014}\natexlab{}.
\newblock \showarticletitle{Automatic Scene Inference for 3D Object
  Compositing}.
\newblock \bibinfo{journal}{\emph{ACM Trans. Graphics}} (\bibinfo{year}{2014}),
  \bibinfo{pages}{32:1--32:15}.
\newblock


\bibitem[Kingma and Ba(2014)]%
        {adam}
\bibfield{author}{\bibinfo{person}{Diederik Kingma} {and}
  \bibinfo{person}{Jimmy Ba}.} \bibinfo{year}{2014}\natexlab{}.
\newblock \showarticletitle{Adam: A method for stochastic optimization}.
\newblock \bibinfo{journal}{\emph{arXiv preprint arXiv:1412.6980}}
  (\bibinfo{year}{2014}).
\newblock


\bibitem[LeGendre et~al\mbox{.}(2019)]%
        {legendre2019deeplight}
\bibfield{author}{\bibinfo{person}{Chloe LeGendre}, \bibinfo{person}{Wan-Chun
  Ma}, \bibinfo{person}{Graham Fyffe}, \bibinfo{person}{John Flynn},
  \bibinfo{person}{Laurent Charbonnel}, \bibinfo{person}{Jay Busch}, {and}
  \bibinfo{person}{Paul Debevec}.} \bibinfo{year}{2019}\natexlab{}.
\newblock \showarticletitle{Deeplight: Learning illumination for unconstrained
  mobile mixed reality}. In \bibinfo{booktitle}{\emph{Proc.~CVPR}}.
  \bibinfo{pages}{5918--5928}.
\newblock


\bibitem[Li et~al\mbox{.}(2018)]%
        {li2018interiornet}
\bibfield{author}{\bibinfo{person}{Wenbin Li}, \bibinfo{person}{Sajad Saeedi},
  \bibinfo{person}{John McCormac}, \bibinfo{person}{Ronald Clark},
  \bibinfo{person}{Dimos Tzoumanikas}, \bibinfo{person}{Qing Ye},
  \bibinfo{person}{Yuzhong Huang}, \bibinfo{person}{Rui Tang}, {and}
  \bibinfo{person}{Stefan Leutenegger}.} \bibinfo{year}{2018}\natexlab{}.
\newblock \showarticletitle{InteriorNet: Mega-scale multi-sensor
  photo-realistic indoor scenes dataset}. In
  \bibinfo{booktitle}{\emph{Proc.~BMVC}}.
\newblock


\bibitem[Li et~al\mbox{.}(2020a)]%
        {li2020inverse}
\bibfield{author}{\bibinfo{person}{Zhengqin Li}, \bibinfo{person}{Mohammad
  Shafiei}, \bibinfo{person}{Ravi Ramamoorthi}, \bibinfo{person}{Kalyan
  Sunkavalli}, {and} \bibinfo{person}{Manmohan Chandraker}.}
  \bibinfo{year}{2020}\natexlab{a}.
\newblock \showarticletitle{Inverse Rendering for Complex Indoor Scenes: Shape,
  Spatially-Varying Lighting and {SVBRDF} from a Single Image}. In
  \bibinfo{booktitle}{\emph{Proc.~CVPR}}.
\newblock


\bibitem[Li et~al\mbox{.}(2022)]%
        {li2022physically}
\bibfield{author}{\bibinfo{person}{Zhengqin Li}, \bibinfo{person}{Jia Shi},
  \bibinfo{person}{Sai Bi}, \bibinfo{person}{Rui Zhu}, \bibinfo{person}{Kalyan
  Sunkavalli}, \bibinfo{person}{Milo{\v{s}} Ha{\v{s}}an},
  \bibinfo{person}{Zexiang Xu}, \bibinfo{person}{Ravi Ramamoorthi}, {and}
  \bibinfo{person}{Manmohan Chandraker}.} \bibinfo{year}{2022}\natexlab{}.
\newblock \showarticletitle{Physically-Based Editing of Indoor Scene Lighting
  from a Single Image}.
\newblock \bibinfo{journal}{\emph{arXiv preprint arXiv:2205.09343}}
  (\bibinfo{year}{2022}).
\newblock


\bibitem[Li et~al\mbox{.}(2020b)]%
        {li2020openrooms}
\bibfield{author}{\bibinfo{person}{Zhengqin Li}, \bibinfo{person}{Ting-Wei Yu},
  \bibinfo{person}{Shen Sang}, \bibinfo{person}{Sarah Wang},
  \bibinfo{person}{Meng Song}, \bibinfo{person}{Yuhan Liu},
  \bibinfo{person}{Yu-Ying Yeh}, \bibinfo{person}{Rui Zhu},
  \bibinfo{person}{Nitesh Gundavarapu}, \bibinfo{person}{Jia Shi},
  {et~al\mbox{.}}} \bibinfo{year}{2020}\natexlab{b}.
\newblock \showarticletitle{OpenRooms: An End-to-End Open Framework for
  Photorealistic Indoor Scene Datasets}.
\newblock \bibinfo{journal}{\emph{Proc.~CVPR}} (\bibinfo{year}{2020}).
\newblock


\bibitem[Liu et~al\mbox{.}(2019)]%
        {liu2019neural}
\bibfield{author}{\bibinfo{person}{Chao Liu}, \bibinfo{person}{Jinwei Gu},
  \bibinfo{person}{Kihwan Kim}, \bibinfo{person}{Srinivasa~G Narasimhan}, {and}
  \bibinfo{person}{Jan Kautz}.} \bibinfo{year}{2019}\natexlab{}.
\newblock \showarticletitle{Neural rgb (r) d sensing: Depth and uncertainty
  from a video camera}. In \bibinfo{booktitle}{\emph{Proceedings of the
  IEEE/CVF Conference on Computer Vision and Pattern Recognition}}.
  \bibinfo{pages}{10986--10995}.
\newblock


\bibitem[Luo et~al\mbox{.}(2020)]%
        {luo2020consistent}
\bibfield{author}{\bibinfo{person}{Xuan Luo}, \bibinfo{person}{Jia-Bin Huang},
  \bibinfo{person}{Richard Szeliski}, \bibinfo{person}{Kevin Matzen}, {and}
  \bibinfo{person}{Johannes Kopf}.} \bibinfo{year}{2020}\natexlab{}.
\newblock \showarticletitle{Consistent video depth estimation}.
\newblock \bibinfo{journal}{\emph{ACM Transactions on Graphics (ToG)}}
  \bibinfo{volume}{39}, \bibinfo{number}{4} (\bibinfo{year}{2020}),
  \bibinfo{pages}{71--1}.
\newblock


\bibitem[McCormac et~al\mbox{.}(2017)]%
        {mccormac2017scenenet}
\bibfield{author}{\bibinfo{person}{John McCormac}, \bibinfo{person}{Ankur
  Handa}, \bibinfo{person}{Stefan Leutenegger}, {and} \bibinfo{person}{Andrew~J
  Davison}.} \bibinfo{year}{2017}\natexlab{}.
\newblock \showarticletitle{Scenenet rgb-d: Can 5m synthetic images beat
  generic imagenet pre-training on indoor segmentation?}. In
  \bibinfo{booktitle}{\emph{Proc.~ICCV}}. \bibinfo{pages}{2678--2687}.
\newblock


\bibitem[Niklaus et~al\mbox{.}(2019)]%
        {niklaus20193d}
\bibfield{author}{\bibinfo{person}{Simon Niklaus}, \bibinfo{person}{Long Mai},
  \bibinfo{person}{Jimei Yang}, {and} \bibinfo{person}{Feng Liu}.}
  \bibinfo{year}{2019}\natexlab{}.
\newblock \showarticletitle{3d ken burns effect from a single image}.
\newblock \bibinfo{journal}{\emph{ACM Transactions on Graphics (ToG)}}
  \bibinfo{volume}{38}, \bibinfo{number}{6} (\bibinfo{year}{2019}),
  \bibinfo{pages}{1--15}.
\newblock


\bibitem[Ramamoorthi and Hanrahan(2001)]%
        {ramamoorthi2001efficient}
\bibfield{author}{\bibinfo{person}{Ravi Ramamoorthi} {and} \bibinfo{person}{Pat
  Hanrahan}.} \bibinfo{year}{2001}\natexlab{}.
\newblock \showarticletitle{An efficient representation for irradiance
  environment maps}. In \bibinfo{booktitle}{\emph{ACM Trans. Graphics}}.
\newblock


\bibitem[Sch\"{o}nberger et~al\mbox{.}(2016)]%
        {schoenberger2016mvs}
\bibfield{author}{\bibinfo{person}{Johannes~Lutz Sch\"{o}nberger},
  \bibinfo{person}{Enliang Zheng}, \bibinfo{person}{Marc Pollefeys}, {and}
  \bibinfo{person}{Jan-Michael Frahm}.} \bibinfo{year}{2016}\natexlab{}.
\newblock \showarticletitle{Pixelwise View Selection for Unstructured
  Multi-View Stereo}. In \bibinfo{booktitle}{\emph{European Conference on
  Computer Vision (ECCV)}}.
\newblock


\bibitem[Sengupta et~al\mbox{.}(2019)]%
        {sengupta2019neural}
\bibfield{author}{\bibinfo{person}{Soumyadip Sengupta}, \bibinfo{person}{Jinwei
  Gu}, \bibinfo{person}{Kihwan Kim}, \bibinfo{person}{Guilin Liu},
  \bibinfo{person}{David~W Jacobs}, {and} \bibinfo{person}{Jan Kautz}.}
  \bibinfo{year}{2019}\natexlab{}.
\newblock \showarticletitle{Neural inverse rendering of an indoor scene from a
  single image}. In \bibinfo{booktitle}{\emph{Proc.~ICCV}}.
\newblock


\bibitem[Silberman et~al\mbox{.}(2012)]%
        {silberman2012indoor}
\bibfield{author}{\bibinfo{person}{Nathan Silberman}, \bibinfo{person}{Derek
  Hoiem}, \bibinfo{person}{Pushmeet Kohli}, {and} \bibinfo{person}{Rob
  Fergus}.} \bibinfo{year}{2012}\natexlab{}.
\newblock \showarticletitle{Indoor segmentation and support inference from rgbd
  images}. In \bibinfo{booktitle}{\emph{Proc.~ECCV}}.
\newblock


\bibitem[Sitzmann et~al\mbox{.}(2019)]%
        {sitzmann2019deepvoxels}
\bibfield{author}{\bibinfo{person}{Vincent Sitzmann}, \bibinfo{person}{Justus
  Thies}, \bibinfo{person}{Felix Heide}, \bibinfo{person}{Matthias
  Nie{\ss}ner}, \bibinfo{person}{Gordon Wetzstein}, {and}
  \bibinfo{person}{Michael Zollhofer}.} \bibinfo{year}{2019}\natexlab{}.
\newblock \showarticletitle{Deepvoxels: Learning persistent 3d feature
  embeddings}. In \bibinfo{booktitle}{\emph{Proceedings of the IEEE/CVF
  Conference on Computer Vision and Pattern Recognition}}.
  \bibinfo{pages}{2437--2446}.
\newblock


\bibitem[Somanath and Kurz(2021)]%
        {somanath2021hdr}
\bibfield{author}{\bibinfo{person}{Gowri Somanath} {and}
  \bibinfo{person}{Daniel Kurz}.} \bibinfo{year}{2021}\natexlab{}.
\newblock \showarticletitle{HDR Environment Map Estimation for Real-Time
  Augmented Reality}. In \bibinfo{booktitle}{\emph{Proceedings of the IEEE/CVF
  Conference on Computer Vision and Pattern Recognition}}.
  \bibinfo{pages}{11298--11306}.
\newblock


\bibitem[Song and Funkhouser(2019)]%
        {Song_2019_CVPR}
\bibfield{author}{\bibinfo{person}{Shuran Song} {and} \bibinfo{person}{Thomas
  Funkhouser}.} \bibinfo{year}{2019}\natexlab{}.
\newblock \showarticletitle{Neural Illumination: Lighting Prediction for Indoor
  Environments}. In \bibinfo{booktitle}{\emph{Proc.~CVPR}}.
  \bibinfo{pages}{6918--6926}.
\newblock


\bibitem[Srinivasan et~al\mbox{.}(2020)]%
        {srinivasan2020lighthouse}
\bibfield{author}{\bibinfo{person}{Pratul~P Srinivasan}, \bibinfo{person}{Ben
  Mildenhall}, \bibinfo{person}{Matthew Tancik}, \bibinfo{person}{Jonathan~T
  Barron}, \bibinfo{person}{Richard Tucker}, {and} \bibinfo{person}{Noah
  Snavely}.} \bibinfo{year}{2020}\natexlab{}.
\newblock \showarticletitle{Lighthouse: Predicting Lighting Volumes for
  Spatially-Coherent Illumination}. In \bibinfo{booktitle}{\emph{Proc.~CVPR}}.
  \bibinfo{pages}{8080--8089}.
\newblock


\bibitem[Wang et~al\mbox{.}(2021)]%
        {wang2021learning}
\bibfield{author}{\bibinfo{person}{Zian Wang}, \bibinfo{person}{Jonah Philion},
  \bibinfo{person}{Sanja Fidler}, {and} \bibinfo{person}{Jan Kautz}.}
  \bibinfo{year}{2021}\natexlab{}.
\newblock \showarticletitle{Learning indoor inverse rendering with 3d
  spatially-varying lighting}. In \bibinfo{booktitle}{\emph{Proceedings of the
  IEEE/CVF International Conference on Computer Vision}}.
  \bibinfo{pages}{12538--12547}.
\newblock


\bibitem[Wei et~al\mbox{.}(2020)]%
        {wei2020object}
\bibfield{author}{\bibinfo{person}{Xin Wei}, \bibinfo{person}{Guojun Chen},
  \bibinfo{person}{Yue Dong}, \bibinfo{person}{Stephen Lin}, {and}
  \bibinfo{person}{Xin Tong}.} \bibinfo{year}{2020}\natexlab{}.
\newblock \showarticletitle{Object-based Illumination Estimation with
  Rendering-aware Neural Networks}.
\newblock \bibinfo{journal}{\emph{Proc.~ECCV}}.
\newblock


\bibitem[Zhan et~al\mbox{.}(2022)]%
        {zhan2022gmlight}
\bibfield{author}{\bibinfo{person}{Fangneng Zhan}, \bibinfo{person}{Yingchen
  Yu}, \bibinfo{person}{Changgong Zhang}, \bibinfo{person}{Rongliang Wu},
  \bibinfo{person}{Wenbo Hu}, \bibinfo{person}{Shijian Lu},
  \bibinfo{person}{Feiying Ma}, \bibinfo{person}{Xuansong Xie}, {and}
  \bibinfo{person}{Ling Shao}.} \bibinfo{year}{2022}\natexlab{}.
\newblock \showarticletitle{Gmlight: Lighting estimation via geometric
  distribution approximation}.
\newblock \bibinfo{journal}{\emph{IEEE Transactions on Image Processing}}
  \bibinfo{volume}{31} (\bibinfo{year}{2022}), \bibinfo{pages}{2268--2278}.
\newblock


\bibitem[Zhan et~al\mbox{.}(2021a)]%
        {zhan2021sparse}
\bibfield{author}{\bibinfo{person}{Fangneng Zhan}, \bibinfo{person}{Changgong
  Zhang}, \bibinfo{person}{Wenbo Hu}, \bibinfo{person}{Shijian Lu},
  \bibinfo{person}{Feiying Ma}, \bibinfo{person}{Xuansong Xie}, {and}
  \bibinfo{person}{Ling Shao}.} \bibinfo{year}{2021}\natexlab{a}.
\newblock \showarticletitle{Sparse needlets for lighting estimation with
  spherical transport loss}. In \bibinfo{booktitle}{\emph{Proceedings of the
  IEEE/CVF International Conference on Computer Vision}}.
  \bibinfo{pages}{12830--12839}.
\newblock


\bibitem[Zhan et~al\mbox{.}(2021b)]%
        {zhan2021emlight}
\bibfield{author}{\bibinfo{person}{Fangneng Zhan}, \bibinfo{person}{Changgong
  Zhang}, \bibinfo{person}{Yingchen Yu}, \bibinfo{person}{Yuan Chang},
  \bibinfo{person}{Shijian Lu}, \bibinfo{person}{Feiying Ma}, {and}
  \bibinfo{person}{Xuansong Xie}.} \bibinfo{year}{2021}\natexlab{b}.
\newblock \showarticletitle{Emlight: Lighting estimation via spherical
  distribution approximation}. In \bibinfo{booktitle}{\emph{Proceedings of the
  AAAI Conference on Artificial Intelligence}}, Vol.~\bibinfo{volume}{35}.
  \bibinfo{pages}{3287--3295}.
\newblock


\bibitem[Zhang et~al\mbox{.}(2021)]%
        {zhang2021consistent}
\bibfield{author}{\bibinfo{person}{Zhoutong Zhang}, \bibinfo{person}{Forrester
  Cole}, \bibinfo{person}{Richard Tucker}, \bibinfo{person}{William~T Freeman},
  {and} \bibinfo{person}{Tali Dekel}.} \bibinfo{year}{2021}\natexlab{}.
\newblock \showarticletitle{Consistent depth of moving objects in video}.
\newblock \bibinfo{journal}{\emph{ACM Transactions on Graphics (TOG)}}
  \bibinfo{volume}{40}, \bibinfo{number}{4} (\bibinfo{year}{2021}),
  \bibinfo{pages}{1--12}.
\newblock


\bibitem[Zhao and Guo(2020)]%
        {zhao2020pointar}
\bibfield{author}{\bibinfo{person}{Yiqin Zhao} {and} \bibinfo{person}{Tian
  Guo}.} \bibinfo{year}{2020}\natexlab{}.
\newblock \showarticletitle{Pointar: Efficient lighting estimation for mobile
  augmented reality}. In \bibinfo{booktitle}{\emph{European Conference on
  Computer Vision}}. Springer, \bibinfo{pages}{678--693}.
\newblock


\bibitem[Zhu et~al\mbox{.}(2022)]%
        {zhu2022irisformer}
\bibfield{author}{\bibinfo{person}{Rui Zhu}, \bibinfo{person}{Zhengqin Li},
  \bibinfo{person}{Janarbek Matai}, \bibinfo{person}{Fatih Porikli}, {and}
  \bibinfo{person}{Manmohan Chandraker}.} \bibinfo{year}{2022}\natexlab{}.
\newblock \showarticletitle{IRISformer: Dense Vision Transformers for
  Single-Image Inverse Rendering in Indoor Scenes}. In
  \bibinfo{booktitle}{\emph{Proceedings of the IEEE/CVF Conference on Computer
  Vision and Pattern Recognition}}. \bibinfo{pages}{2822--2831}.
\newblock


\end{thebibliography}
